\documentclass[lettersize,journal]{IEEEtran}
\usepackage{amsmath,amsfonts}
\usepackage{algorithm}
\usepackage{subcaption}
\usepackage{array}
\usepackage{textcomp}
\usepackage{stfloats}
\usepackage{url}
\usepackage{verbatim}
\usepackage{graphicx}
\usepackage{cite}
\usepackage{bm}
\usepackage{xcolor}
\usepackage{lettrine}
\usepackage{booktabs}
\usepackage{diagbox}
\usepackage{mathtools}
\usepackage{colortbl}
\usepackage{color}
\usepackage{tikz}
\usepackage{bbding}
\usepackage{amssymb}
\usepackage{multirow}
\usepackage{xcolor} 
\usepackage{subcaption}
\usepackage{threeparttable}
\usepackage{arydshln} 
\usepackage{booktabs}   
\usepackage{siunitx}
\usepackage{amsthm}

\theoremstyle{definition}

\theoremstyle{plain}

\usepackage{enumitem}
\usepackage{tabu}   
\usepackage[colorlinks=true, citecolor=blue]{hyperref}
\usepackage{algpseudocode}

\definecolor{myred}{RGB}{219, 33, 38}
\definecolor{grayblue}{RGB}{222, 228, 238}

\begin{document}

\title{BayesTTA: Continual-Temporal Test-Time Adaptation for Vision-Language Models via Gaussian Discriminant Analysis}

\author{{Shuang~Cui, Jinglin~Xu, Yi~Li, Xiongxin~Tang, Jiangmeng~Li, Jiahuan~Zhou, Fanjiang~Xu, Fuchun~Sun, \IEEEmembership{Fellow,~IEEE} and Hui~Xiong, \IEEEmembership{Fellow,~IEEE}}

%,  \IEEEmembership{Staff,~IEEE,
% <-this % stops a space
%\thanks{This paper was produced by the IEEE Publication Technology Group. They are in Piscataway, NJ.}% <-this % stops a space
%\thanks{Manuscript received April 19, 2021; revised August 16, 2021.}
\thanks{Shuang Cui, Jinglin Xu, Yi Li, Xiongxin Tang, Jiangmeng Li, and Fanjiang Xu are with the National Key Laboratory of Space Integrated Information System, Institute of Software Chinese Academy of Sciences, and University of Chinese Academy of Sciences, Beijing, China; E-mail: \{cuishuang21, xujinglin23, liyi212\}@mails.ucas.ac.cn; \{xiongxin, jiangmeng2019, fanjiang\}@iscas.ac.cn.

Jiahuan Zhou is with the Wangxuan Institute of Computer Technology, Peking University, Beijing, China. E-mail: jiahuanzhou@pku.edu.cn.

Fuchun Sun is with the Department of Computer Science and Technology, Tsinghua University, Beijing, China. E-mail: fcsun@mail.tsinghua.edu.cn.

Hui Xiong is with Thrust of Artificial Intelligence, the Hong Kong University of Science and Technology (Guangzhou), Guangzhou, China. He is also with Department of Computer Science \& Engineering, the Hong Kong University of Science and Technology, Hong Kong SAR, China. E-mail: xionghui@ust.hk.

(Corresponding author: Jiangmeng Li)}}

% The paper headers
\markboth{Journal of \LaTeX\ Class Files,~Vol.~14, No.~8, August~2021}%
{Shell \MakeLowercase{\textit{et al.}}: A Sample Article Using IEEEtran.cls for IEEE Journals}

%\IEEEpubid{0000--0000/00\$00.00 \copyright~2021 IEEE}
% Remember, if you use this you must call \IEEEpubidadjcol in the second
% column for its text to clear the IEEEpubid mark.

\maketitle

\begin{abstract}
Vision-language models (VLMs) such as CLIP achieve strong zero-shot recognition but degrade significantly under \textit{temporally evolving distribution shifts} common in real-world scenarios (e.g., gradual illumination or seasonal changes). Existing continual test-time adaptation (CTTA) methods are typically built around sudden and severe distribution shifts and neglect temporal continuity, leading to three core defects: limited memory cache restricts long-range distribution modeling, causing catastrophic forgetting; entropy-based confidence becomes unreliable under temporal drift, worsening error accumulation; and static visual representations misalign with evolving inputs. We formalize this practical problem as \textit{Continual-Temporal Test-Time Adaptation (CT-TTA)}, where test distributions evolve gradually over time. To address it, we propose \textit{BayesTTA}, a Bayesian adaptation framework that enforces temporally consistent predictions and dynamically aligns visual representations. Specifically, BayesTTA incrementally estimates class-conditional Gaussian mixture distributions without storing raw data, adaptively selects covariance structures through statistical hypothesis testing, and performs calibrated inference using Gaussian discriminant analysis (GDA). These calibrated predictions supervise self-paced adaptation of normalization layers, ensuring efficient and stable representation alignment. We establish a comprehensive CT-TTA benchmark across four temporally evolving datasets and further evaluate generalization on ten standard TTA datasets. Extensive experiments show that BayesTTA consistently outperforms state-of-the-art methods, achieving significant gains while maintaining efficiency. Code is available at \href{https://github.com/cuishuang99/BayesTTA}{https://github.com/cuishuang99/BayesTTA}.
\end{abstract}

\begin{IEEEkeywords}
Vision-Language Models, Continual Test-Time Adaptation, Gaussian Discriminant Analysis, Statistical Hypothesis Testing. 
\end{IEEEkeywords}

\section{Introduction} 
\label{sec:introduction}
\setlength{\parskip}{0pt}

\begin{figure}
    \centering
    \includegraphics[width=0.46\textwidth]{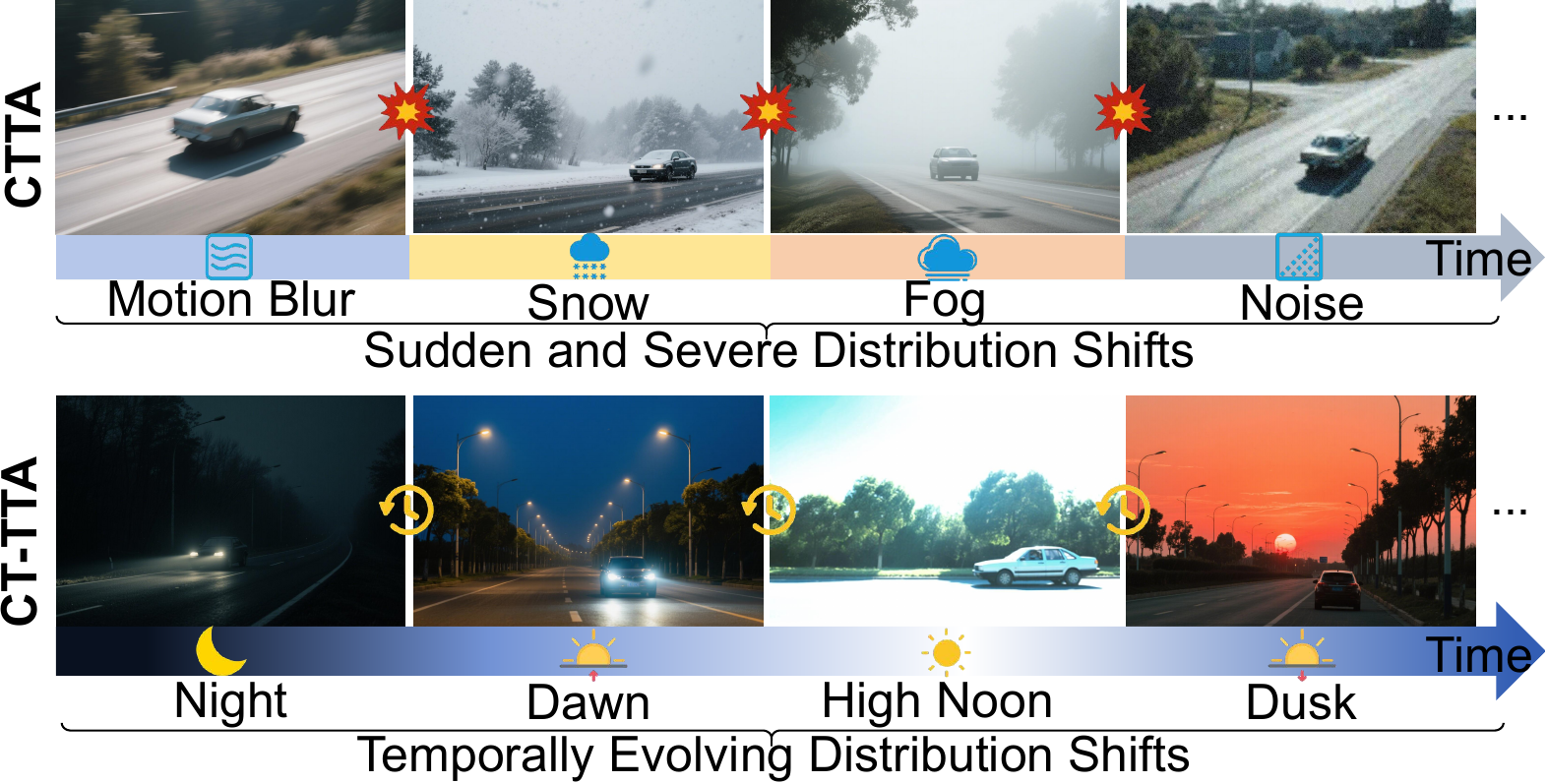}
    \caption{Test-time adaptation paradigms: conventional CTTA (top) handles sudden and severe distribution shifts; CT-TTA (bottom) models temporally evolving distribution shifts.}
    \label{fig:setting}
\end{figure}

\begin{figure}
    \centering
    \includegraphics[width=0.44\textwidth]{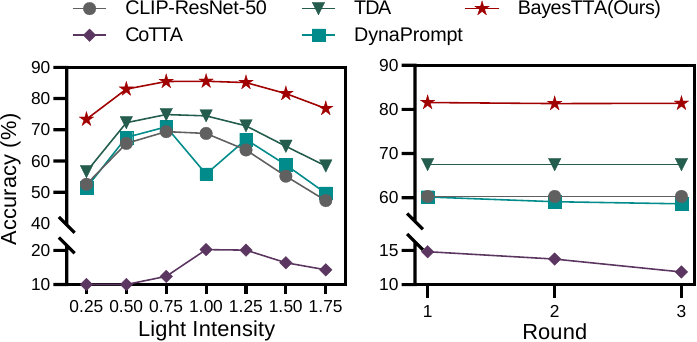}
    \caption{Empirical motivation for CT-TTA. Left: per-domain performance under gradual illumination shifts; Right: averaged per-round performance over all domains.}
    \label{fig:motivation-result}
\end{figure}

\lettrine{L}ARGE-SCALE vision-language models (VLMs), such as CLIP \cite{radford2021learning} and ALIGN \cite{jia2021scaling}, exhibit strong generalization by aligning visual and textual modalities within a shared semantic space. Trained on massive image-text paired datasets \cite{schuhmann2021laion, schuhmann2022laion}, these models enable zero-shot recognition across diverse downstream tasks without task-specific supervision. However, their performance degrades substantially under distribution shifts frequently encountered in real-world deployment, particularly when test data gradually diverge from the training distribution \cite{zhou2022learning,zhang2024vision}. To mitigate this degradation, continual test-time adaptation (CTTA) has emerged as a promising solution \cite{wang2021tent, wang2022continual, niu2023towards, yuan2023robust, tan2025uncertainty, liang2025comprehensive}, enabling online adaptation to unlabeled, non-stationary test streams. While recent CTTA methods have been adapted to VLMs \cite{zhang2024historical, xiao2025dynaprompt}, they predominantly address \textit{sudden and severe} distribution shifts. For example, the widely used CIFAR10-C benchmark \cite{hendrycksbenchmarking2019} contains ten distinct domain styles such as blur, snow, fog, or noise (Fig. \ref{fig:setting}). Nevertheless, this setup may inadequately represent all real-world applications that involve gradual and temporally correlated distribution shifts. For instance, autonomous vehicles encounter gradual illumination changes throughout the day; satellite imagery varies seasonally; industrial sensors drift due to hardware aging \cite{vergara2012chemical}; and weather systems exhibit temporally correlated dynamics. These \textit{temporally evolving distribution shifts} \cite{hoffman2014continuous, yao2022wild, xie2023evolving} exhibit underlying evolutionary patterns \cite{zeng2023foresee}, but existing CTTA methods cannot adequately model such patterns. As a result, their performance remains suboptimal under such scenarios.

To better characterize this setting, we introduce \textbf{Continual-Temporal Test-Time Adaptation (CT-TTA)}, a realistic formulation that explicitly models temporally correlated, gradually evolving distribution shifts. Unlike conventional CTTA, CT-TTA captures temporal dynamics to support more stable adaptation in real-world data streams (Fig. \ref{fig:setting}). To empirically illustrate its necessity, we construct \textbf{CIFAR10-Light}, a synthetic dataset simulating smooth illumination transitions via Fourier-based amplitude scaling \cite{brigham1988fast} on CIFAR-10 \cite{krizhevsky2009learning}. As shown in Fig. \ref{fig:motivation-result}, CLIP's zero-shot accuracy drops from 68.83\% under standard illumination to 47.33\% under overexposure, revealing its vulnerability to temporal drift. State-of-the-art (SOTA) CTTA methods such as CoTTA \cite{wang2022continual} and DynaPrompt \cite{xiao2025dynaprompt} suffer similarly, reflecting their inability to model temporal evolutionary patterns. This degradation arises from three fundamental limitations of existing CTTA methods under temporally evolving distribution shifts: 1) limited cache capacity impedes long-range distribution modeling, leading to catastrophic forgetting as historical information is gradually overwritten; 2) entropy-based confidence estimation becomes increasingly unreliable under continuous temporal drift, compounding pseudo-label errors and exacerbating error accumulation; and 3) static visual representations remain misaligned with temporally evolving inputs, intensifying representation drift and destabilizing the adaptation process.

To address these limitations, we propose \textbf{BayesTTA}, a principled Bayesian framework for VLMs in CT-TTA that enforces temporally consistent predictions and dynamically aligns visual representations. BayesTTA mitigates catastrophic forgetting by incrementally estimating class-conditional Gaussian mixture distributions via a \textit{debiased temporal distribution estimation} strategy. This allows long-range modeling of non-stationary distributions without retaining raw data, thereby preserving historical knowledge under memory constraints \cite{karmanov2024efficient,zhang2024historical, xiao2025dynaprompt}. To combat pseudo-label degradation caused by unreliable entropy-based confidence under continuous drift, BayesTTA performs temporally consistent Bayesian inference via Gaussian discriminant analysis (GDA) \cite{hastie2009elements}. By modeling class-conditional densities with estimated means and covariances, GDA provides a statistically grounded and analytically tractable prediction rule. Given the approximately Gaussian nature of features in high-dimensional embedding spaces \cite{bishop2006pattern}, this approach naturally fits the evolving test-time setting and yields more reliable calibration. To further improve the accuracy of distribution estimation and calibration, BayesTTA integrates a \textit{covariance homogeneity test} based on Principal Component Analysis (PCA) \cite{jolliffe2002principal} and Box's $M$ test \cite{box1949general}. This test enables dynamic selection between \textit{homogeneous} and \textit{heterogeneous} covariance structures, avoiding the restrictive identity covariance assumption adopted by prior methods \cite{wang2024a, han2024dota, dai2025free} and enhancing adaptation flexibility. Finally, to mitigate representation drift, BayesTTA online adapts the image encoder by leveraging fused pseudo-labels from GDA-based predictions and CLIP's zero-shot outputs. The adaptation is applied selectively to normalization layers for efficiency and further stabilized via an exponential moving average (EMA) across time, ensuring smooth and robust representation alignment. Overall, BayesTTA provides a unified and lightweight solution to error accumulation, catastrophic forgetting, and representation misalignment under realistic CT-TTA conditions.

To evaluate BayesTTA, we construct a \textbf{comprehensive CT-TTA benchmark} covering both real-world and synthetic datasets, including fMoW, Yearbook \cite{yao2022wild}, RMNIST, and CIFAR10-Light. We further evaluate generalization on the standard TTA benchmark across ten cross-domain datasets. Extensive experiments demonstrate that BayesTTA achieves superior performance and efficiency compared to SOTA methods. Our main contributions can be summarized as follows:
\begin{itemize}
\item We formulate Continual-Temporal Test-Time Adaptation (CT-TTA), a practical paradigm that accounts for temporally evolving distribution shifts in real-world VLMs deployment, revealing the limitations of conventional CTTA methods under such conditions.
\item We propose BayesTTA, a Bayesian framework for CT-TTA that explicitly models evolving distributions, adaptively selects covariance structures through statistical hypothesis testing, performs temporally consistent Bayesian inference for calibrated predictions, and dynamically aligns visual representations with evolving inputs.
\item We establish a comprehensive CT-TTA benchmark across four temporally evolving datasets and evaluate generalization on the standard TTA benchmark. Extensive experiments show that BayesTTA achieves stable and significant improvements, surpassing SOTA methods by +14.27\% on RMNIST (ViT-B/16) under CT-TTA.
\end{itemize}

\section{Related Work}
\label{sec:related}
This section reviews recent advances in vision-language models, continual test-time adaptation, and evolving/temporal domain generalization.

\subsection{Vision-Language Models}
\label{subsection:vision-language-models}
Vision-language models (VLMs) achieve strong generalization by aligning visual and textual modalities through large-scale contrastive pretraining on image-text pairs. The seminal CLIP model \cite{radford2021learning} learns a shared embedding space via a dual-encoder design, enabling robust zero-shot transfer to diverse downstream tasks such as open-world segmentation \cite{wang2022cris,zhou2023zegclip} and action recognition \cite{wang2023actionclip,wang2024clip}. Recent efforts to enhance VLMs performance typically follow two directions: (1) prompt engineering to optimize textual inputs, as in CoOp \cite{zhou2022learning}, CoCoOp \cite{zhou2022conditional} and Maple \cite{khattak2023maple}; and (2) visual feature adaptation to refine representations, exemplified by CLIP-Adapter \cite{gao2024clip} and Tip-Adapter \cite{zhang2022tip}. Some methods further exploit feature statistics for training-free adaptation \cite{wang2024a}, showing promise in low-resource settings. However, most existing techniques rely on labeled target data, which limits their applicability in real-world scenarios with scarce annotations or privacy constraints. In this context, test-time adaptation (TTA) \cite{wang2021tent} emerges as a compelling alternative by enabling online model adaptation using only unlabeled target data, without requiring source data or supervision. Building upon this paradigm, we propose a principled TTA framework specifically designed to address the challenges posed by temporally evolving distributions in practical VLMs applications.

\subsection{Continual Test-Time Adaptation}
\label{subsection:continual-test-time-adaptation}
Test-time adaptation (TTA) \cite{wang2021tent,zhao2022delta,zhang2022memo,niu2022efficient,zhang2023domainadaptor,roy2023test,lee2024entropy} enables models to adapt online during inference, addressing distribution shifts between training and test data without requiring access to source data. The representative method, TENT \cite{wang2021tent}, minimizes prediction entropy by adjusting batch normalization statistics for test-time feature alignment. Extending this paradigm to VLMs, TPT \cite{shu2022test} performs gradient-based optimization of learnable text prompts, also based on entropy minimization. To improve the robustness of TPT, Adaprompt \cite{zhang2024robust} introduces handcrafted prompt ensembles and a confidence-aware buffer, while SCP \cite{wang2024towards} enhances stability through a self-text distillation strategy based on conjugate pseudo-labels. WATT \cite{osowiechi2024watt} employs ensemble learning through weight averaging over diverse text prompts. Despite their effectiveness, most approaches rely on backpropagation and heavy data augmentation, leading to significant computational overhead at test time. To reduce this cost, TDA \cite{karmanov2024efficient} proposes a training-free dynamic adapter that caches representative test-time features, and DPE \cite{zhang2024dual} introduces dual prototype sets to aggregate multimodal task knowledge. BCA \cite{zhou2025bayesian} leverages Bayesian updating of class priors based on test-time posteriors but remains confined within a discriminative modeling framework without explicit feature distribution modeling. Notably, most existing TTA methods assume a stationary target distribution.

Continual test-time adaptation (CTTA) \cite{wang2022continual,niu2023towards,yuan2023robust,dobler2023robust,gan2023decorate,song2023ecotta,liu2024vida,zhang2024dynamic} extends conventional TTA to continually changing target domains. However, its unsupervised and continual nature exacerbates two major challenges: error accumulation and catastrophic forgetting. CoTTA \cite{wang2022continual} mitigates error accumulation by refining pseudo-labels through weight-averaged and augmentation-averaged predictions. SAR \cite{niu2023towards} improves stability under dynamic shifts through sharpness-aware entropy minimization. RoTTA \cite{yuan2023robust} improves robustness to temporal shifts via memory replay and robust normalization but does not model long-range temporal evolution, limiting sustained adaptation. For VLM-specific scenarios, HisTPT \cite{zhang2024historical} employs multiple knowledge banks to reduce forgetting, while DynaPrompt \cite{xiao2025dynaprompt} introduces a dynamic prompt buffer to adaptively exploit informative samples while curbing noise propagation. Similar to TDA \cite{karmanov2024efficient}, these approaches \cite{zhang2024historical, xiao2025dynaprompt} are constrained by limited cache capacity, which restricts the retention of long-term historical information. DOTA \cite{han2024dota} employs GDA under an identical covariance assumption for distribution estimation, which may not hold universally. Moreover, most CTTA methods often assume sudden and severe distribution shifts—such as transitions between distinct corruption types in CIFAR10-C \cite{hendrycksbenchmarking2019}—and treat each domain or batch independently, overlooking the temporal continuity of real-world environments. Actually, real-world distribution shifts typically evolve gradually over time (e.g., illumination changes, seasonal patterns, or sensor drift), making existing CTTA methods suboptimal in such settings. 

To address these limitations, we propose a more realistic formulation: Continual-Temporal Test-Time Adaptation (CT-TTA), which explicitly models temporally correlated, gradually evolving distribution shifts. Under this paradigm, we present BayesTTA, a principled Bayesian framework tailored for VLM adaptation. Unlike prior GDA-based methods \cite{han2024dota,dai2025free}, BayesTTA incorporates a statistical covariance homogeneity test—combining PCA and Box's $M$ test—to adaptively select between homogeneous and heterogeneous covariance structures, enhancing flexibility across varying distributional forms. Additionally, BayesTTA introduces a GDA-based sketch refinement mechanism that uses calibrated predictions to supervise test-time updates of normalization layers, thereby aligning visual representations more effectively with evolving inputs.

\subsection{Evolving/Temporal Domain Generalization}
Recent advances in evolving domain generalization (EDG) and temporal domain generalization (TDG) tackle the challenge of adapting models to gradually shifting domains over time \cite{nasery2021training,bai2023temporal,qin2022generalizing,xie2024enhancing,zeng2024latent}. Existing methods can be broadly categorized into two paradigms. The first explicitly encodes temporal dynamics into model architectures. GI \cite{nasery2021training} extrapolates future domain states via a first-order Taylor expansion of activation functions, but this shallow modeling restricts its expressiveness. In contrast, DRAIN \cite{bai2023temporal} adopts a Bayesian framework with recurrent graph structures to thoroughly capture concept drift and domain evolution. The second line of work seeks to model the latent factors underlying domain shifts. LSSAE \cite{qin2022generalizing} disentangles invariant and dynamic components via variational inference, while MISTS \cite{xie2024enhancing} further introduces mutual information constraints to encourage their complementarity. SDE-EDG \cite{zeng2024latent} advances this direction by modeling continuous domain trajectories through stochastic differential equations, improving temporal resolution. 

However, EDG/TDG methods assume access to multiple source domains during training, which imposes significant computational overhead and limits scalability in resource-constrained environments. EvoS \cite{xie2023evolving} addresses this by operating in a domain-incremental setting, modeling Gaussian feature dynamics, and employing attention mechanisms to capture temporal patterns in evolving feature statistics. W-Diff \cite{xie2024weight} mitigates forgetting and error propagation by modeling classifier dynamics via conditional diffusion in the weight space. In contrast, our method eliminates the need for source data and multi-phase training, enabling scalable and efficient adaptation solely at test time, with strong practical utility in evolving real-world environments.

\section{Preliminaries}
\label{sec:preliminaries}
This section outlines the foundation of CLIP and its adaptation under the continual test-time adaptation (CTTA) setting.

\subsection{Zero-shot CLIP}
\label{subsection:clip}

We adopt the CLIP model \cite{radford2021learning} as the representative VLM, which comprises an image encoder $\boldsymbol{f}_v$ and a text encoder $\boldsymbol{f}_t$, jointly pre-trained on 400 million image-text pairs \cite{schuhmann2021laion} via a contrastive objective that aligns visual and textual representations in a shared embedding space. Given a class label, the text encoder $\boldsymbol{f}_t$ transforms a prompt-augmented class name (e.g., ``a photo of a \{class\}'') into a class prototype $\boldsymbol{w}_k \in \mathbb{R}^D$, where $D$ is the embedding dimension. For a test image $\boldsymbol{x}$, the image encoder $\boldsymbol{f}_v$ produces a visual embedding $\boldsymbol{z} {=} \boldsymbol{f}_v(\boldsymbol{x}) \in \mathbb{R}^D$. Zero-shot classification proceeds by computing cosine similarities between $\boldsymbol{z}$ and all class prototypes, followed by a temperature-scaled softmax:
\begin{equation}
\mathrm{P}(Y = k \mid \boldsymbol{x}) = \frac{\exp\left( \cos(\boldsymbol{z}, \boldsymbol{w}_k) / \tau \right)}{\sum_{j=1}^K \exp\left( \cos(\boldsymbol{z}, \boldsymbol{w}_j) / \tau \right)},
\end{equation}
where $K$ denotes the number of classes, and $\tau$ is a temperature parameter set to 0.01, following prior work.

\subsection{Continual Test-Time Adaptation for CLIP}
\label{subsection:pre_ctta_clip}
The CLIP model is trained on the source domain $\mathcal{P}_s$. CTTA aims to adapt the model during test time to continuously changing target distributions $\mathcal{P}_t$, with $\mathcal{P}_t \neq \mathcal{P}_s$. In this fully test-time setting, adaptation is conducted online without access to source data. At each time step $t$, the model observes only an unlabeled batch $\boldsymbol{X}^t$ and updates parameters immediately.

\section{Methodology}
\label{sec:method}
This section starts by formalizing the CT-TTA problem and then presents an overview of the BayesTTA framework. Detailed descriptions of each BayesTTA component are provided in the following subsections.

\subsection{Continual-Temporal Test-Time Adaptation for Vision-Language Models}
\label{subsection:problem-setting}
We propose \textbf{Continual-Temporal Test-Time Adaptation (CT-TTA)} for VLMs, which explicitly models temporally evolving target distributions encountered during deployment. Specifically, at each time step $ t \in \{1, \ldots, T\} $, the CLIP model receives an unlabeled batch of test samples $\boldsymbol{X}^t {=} \{\boldsymbol{x}_i^t\}_{i=1}^{N_t}$, where each sample is independently drawn from the current target distribution $\mathcal{P}_t$. $N_t$ denotes the number of samples in the current batch. The distribution $\mathcal{P}_t$ evolves gradually over time, subject to a \textit{bounded drift} constraint measured by the Kullback-Leibler (KL) divergence \cite{kullback1997information}:
\begin{equation}
    \mathbb{E}\left[ D_{\mathrm{KL}}(\mathcal{P}_{t} \,\|\, \mathcal{P}_{t+1}) \right] \leq \delta,
\end{equation}
where $\delta > 0$ is a drift threshold.

Following a strict online causality principle, adaptation at time $t$ relies solely on the current batch $\boldsymbol{X}^t$ and the previous model parameters $\boldsymbol{\theta}_{t-1}$, with no access to past data batches or source domain samples. The model updates its parameters through a self-supervised operator $\mathcal{A}$:
\begin{equation}
    \boldsymbol{\theta}_t = \mathcal{A}(\boldsymbol{\theta}_{t-1}, \boldsymbol{X}^t).
\end{equation}

By modeling temporally smooth and bounded transitions, CT-TTA enables continual adaptation to temporally evolving distribution shifts. This setting fundamentally differs from conventional CTTA, which typically assumes abrupt and temporally uncorrelated distribution shifts \cite{wang2022continual,zhang2024historical, xiao2025dynaprompt}.

\begin{figure*}
      \centering        \includegraphics[width=0.8\textwidth]{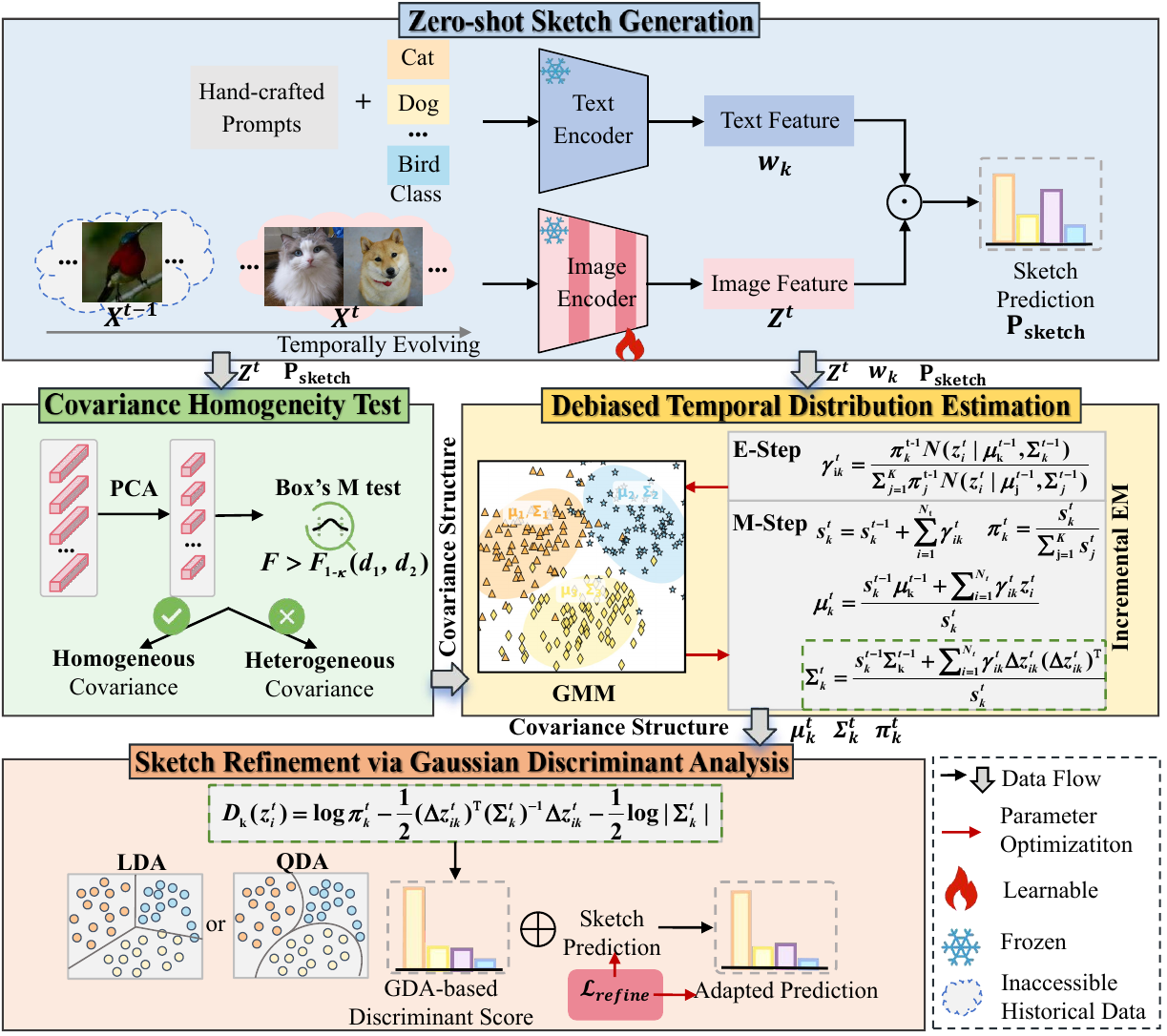}
	\caption{Overview of BayesTTA: (1) Zero-shot Sketch Generation, (2) Covariance Homogeneity Test, (3) Debiased Temporal Distribution Estimation, (4) Sketch Refinement via Gaussian Discriminant Analysis. Only normalization layers (LayerNorm in ViT or BatchNorm in ResNet) of the image encoder are updated, and all other parameters remain frozen during adaptation.}
	\label{fig:frame-cttta}
\end{figure*}

\subsection{Overview}
\label{subsection:overview}
Conventional CTTA methods exhibit three key limitations under temporally evolving distribution shifts in CT-TTA scenarios: error accumulation, catastrophic forgetting, and representation drift. To address these challenges, we propose BayesTTA, a principled Bayesian adaptation framework specifically designed for VLMs in CT-TTA. BayesTTA is based on the modeling assumption that visual embeddings extracted by CLIP follow class-conditional Gaussian mixture distributions, aligning with prior work that models deep features as Gaussian mixtures \cite{bishop2006pattern,wang2024a}. Formally, the class-conditional likelihood is defined as:
\begin{equation}
p_t(\boldsymbol{z} \mid Y = k) = \mathcal{N}(\boldsymbol{\mu}_k^t, \boldsymbol{\Sigma}_k^t),
\end{equation}
where $\boldsymbol{z}$ denotes the visual embedding, and $\boldsymbol{\mu}_k^t$, $\boldsymbol{\Sigma}_k^t$ represent the mean and covariance of class $k$ at time $t$. Given class priors $\pi_k^t = \mathrm{P}_t(Y = k)$, the posterior is computed via Bayes' theorem \cite{bishop2006pattern}:
\begin{equation}
\begin{split}
\mathrm{P}_t(Y = k \mid \boldsymbol{z}_i^t) 
= \frac{\mathcal{N}(\boldsymbol{z}_i^t \mid \boldsymbol{\mu}_k^t, \boldsymbol{\Sigma}_k^t) \cdot \pi_k^t}{\sum_{j=1}^K \mathcal{N}(\boldsymbol{z}_i^t \mid \boldsymbol{\mu}_j^t, \boldsymbol{\Sigma}_j^t) \cdot \pi_j^t}.
\end{split}
\end{equation}

By statistically testing whether class-wise covariance matrices are homogeneous or heterogeneous, the posterior simplifies to a softmax over a linear or quadratic function of $\boldsymbol{z}_i^t$, parameterized by the estimated Gaussian statistics. This formulation enables BayesTTA to perform temporally consistent and distribution-aware adaptation via posterior inference, without relying on heuristic confidence scores or memory replay mechanisms \cite{karmanov2024efficient, xiao2025dynaprompt}. 

As depicted in Fig. \ref{fig:frame-cttta}, BayesTTA integrates four complementary components:
\begin{enumerate}[label=(\arabic*)]
\item \textit{Zero-Shot Sketch Generation} module leverages CLIP to produce probabilistic class priors as semantic guidance;
\item \textit{Covariance Homogeneity Test} module adaptively selects between homogeneous and heterogeneous covariance modeling, improving flexibility and accuracy of posterior inference;
\item \textit{Debiased Temporal Distribution Estimation} module estimates Gaussian statistics via incremental EM, preserving long-range information without storing past samples;
\item \textit{Sketch Refinement via GDA} module fuses semantic priors with calibrated posteriors to generate robust pseudo-labels, which dynamically guide representation alignment via selective updates.
\end{enumerate}
Together, these components form a closed-loop Bayesian adaptation pipeline, enabling BayesTTA to maintain stability and robustness under temporally evolving distribution shifts.

\subsection{Zero-shot Sketch Generation}
\label{subsection:sketch-generation}
BayesTTA initiates adaptation with a zero-shot \textit{sketch}, a coarse but semantically meaningful class probability estimate generated by the pre-trained CLIP model. Leveraging CLIP's strong zero-shot transfer capability, this sketch provides a reliable prior to bootstrap test-time adaptation under distribution shifts without requiring labels. At time step $t$, given an unlabeled input batch $\boldsymbol{X}^t {=} \{\boldsymbol{x}_i^t\}_{i=1}^{N_t}$, the CLIP image encoder extracts visual embeddings $\boldsymbol{Z}^t {=} \{\boldsymbol{z}_i^t\}_{i=1}^{N_t}$. Each class $k$ is represented by a corresponding CLIP text embedding $\boldsymbol{w}_k$. The sketch logits are computed as the cosine similarity:
\begin{equation}
\ell_{k}^{\mathrm{sketch}}(\boldsymbol{z}_i^t) = \cos(\boldsymbol{z}_i^t, \boldsymbol{w}_k),
\label{eq:sketch_logits}
\end{equation}
and transformed into sketch probabilities using softmax:
\begin{equation}
\mathrm{P}_{\mathrm{sketch}}(Y = k \mid \boldsymbol{x}_i^t) 
= \frac{\exp\left( \ell_{k}^{\mathrm{sketch}}(\boldsymbol{z}_i^t) / \tau \right)}{\sum_{j=1}^K \exp\left( \ell_{j}^{\mathrm{sketch}}(\boldsymbol{z}_i^t) / \tau \right)}.
\label{eq:sketch_prob}
\end{equation}

These initial sketch probabilities, although potentially imperfect under distribution shifts, preserve transferable class semantics. They serve as the initialization for estimating distributions (\textbf{Section} \ref{subsection:homogeneity_test}, \ref{subsection:distribution_estimation}) and enable refinement via GDA (\textbf{Section} \ref{subsection:sketch_refinement}), forming the semantic backbone of BayesTTA.

\subsection{Covariance Homogeneity Test}
\label{subsection:homogeneity_test}
\vspace{-0.3\baselineskip}
To support adaptive and principled distribution modeling, BayesTTA conducts a covariance homogeneity test to determine whether class-conditional features share a common covariance structure. In contrast to prior methods that make fixed assumptions of identical class covariances \cite{wang2024a, han2024dota, dai2025free}, BayesTTA formulates this decision as a statistical hypothesis test: the null hypothesis $\mathcal{H}_0$ assumes \textit{homogeneous covariance} (i.e., all classes share a common covariance matrix), while the alternative hypothesis $\mathcal{H}_1$ assumes \textit{heterogeneous covariance}, indicating statistically significant differences across class-conditional covariances. This flexible formulation allows BayesTTA to align its generative assumptions with the empirical structure of incoming test data, thereby enhancing adaptation robustness and generality.

Since testing covariance structures directly in high-dimensional or limited-sample regimes can be unstable, we first apply Principal Component Analysis (PCA) \cite{jolliffe2002principal} to project visual embeddings into a low-dimensional, decorrelated subspace. This transformation reduces estimation variance and ensures that class-specific covariance matrices are positive definite and invertible—conditions necessary for applying Box's $M$ test \cite{box1949general}. Let $\hat{\boldsymbol{z}}_i^t \in \mathbb{R}^d$ denote the PCA-transformed visual feature, where we retain the top $d{=}10$ components. Based on sketch predictions, we estimate class-conditional statistics:
\begin{align}
\hat{n}_k^t &= \sum_{\boldsymbol{x}_i^t\in\boldsymbol{X}^t} \mathrm{P}_{\mathrm{sketch}}(Y=k \mid \boldsymbol{x}_i^t), \nonumber \\
\hat{\boldsymbol{\mu}}_k^t &= \frac{1}{\hat{n}_k^t} \sum_{\boldsymbol{x}_i^t\in\boldsymbol{X}^t} \mathrm{P}_{\mathrm{sketch}}(Y=k \mid \boldsymbol{x}_i^t) \cdot \hat{\boldsymbol{z}}_i^t,  \\
\hat{\boldsymbol{\Sigma}}_k^t &= \frac{1}{\hat{n}_k^t} \sum_{\boldsymbol{x}_i^t\in\boldsymbol{X}^t} \mathrm{P}_{\mathrm{sketch}}(Y=k \mid \boldsymbol{x}_i^t) (\hat{\boldsymbol{z}}_i^t - \hat{\boldsymbol{\mu}}_k^t)(\hat{\boldsymbol{z}}_i^t - \hat{\boldsymbol{\mu}}_k^t)^\top.  \nonumber
\end{align}

Under the null hypothesis $\mathcal{H}_0$ of homogeneous covariances, a pooled covariance estimate is computed across classes:
\begin{equation}
\hat{\boldsymbol{\Sigma}}_{\text{pooled}}^t = \frac{1}{\sum_{k=1}^{K}(\hat{n}_k^t - 1)} \sum_{k=1}^{K} (\hat{n}_k^t - 1) \cdot \hat{\boldsymbol{\Sigma}}_k^t.
\end{equation}

We adopt a regularized form of Box's $M$ test to quantify covariance divergence:
\begin{equation}
M = \sum_{k=1}^{K} (\hat{n}_k^t - 1) \ln |\hat{\boldsymbol{\Sigma}}_{\text{pooled}}^t| - \sum_{k=1}^{K} (\hat{n}_k^t - 1) \ln |\hat{\boldsymbol{\Sigma}}_k^t|.
\end{equation}

To improve statistical robustness under high-dimensional, small-sample conditions, the classical $\chi^2$ approximation is replaced with an $F$-distribution-based correction \cite{geisser1958extension}. The adjusted statistic is:
\begin{equation}
    M^* = \frac{M}{c},
\end{equation}
with scaling factor:
\begin{equation}
    c = 1 - \frac{2d^2 + 3d - 1}{6(d + 1)(K - 1)} \left( \sum_{k=1}^K \frac{1}{\hat{n}_k^t - 1} - \frac{1}{\sum_{k=1}^K \hat{n}_k^t - K} \right). \notag
\end{equation}

We further compute the auxiliary adjustment $\lambda$ and degrees of freedom $(d_1, d_2)$ as:
\begin{align}
\lambda &= \frac{d(d + 1)(K - 1)(K + 1)}{6 \left( \sum_{k=1}^{K} \hat{n}_k^t - K - (K - 1) \right)}, \\
d_1 &= \frac{d(d + 1)(K - 1)}{2}, \quad d_2 = \frac{d_1 + 2}{\lambda + \varepsilon},
\end{align}
where $\varepsilon$ is a small constant added to prevent division by zero. The final F-statistic is approximated as:
\begin{equation}
F = \frac{M^*}{d_1 \left(1 + \frac{M^*}{d_2} \right)}.
\end{equation}

At a significance level of $\kappa {=} 0.05$, the null hypothesis is rejected if $F {>} F_{1-\kappa}(d_1, d_2)$, indicating statistically significant differences in class-conditional covariances. In this case, a heterogeneous covariance structure is adopted; otherwise, the model defaults to a homogeneous assumption. This statistically grounded model selection procedure enables BayesTTA to align its inductive bias with the empirical distributional characteristics of the test data, supporting principled and flexible generative modeling across diverse scenarios.

\subsection{Debiased Temporal Distribution Estimation} 
\label{subsection:distribution_estimation}

\begin{algorithm}[t]
\caption{Debiased Temporal Distribution Estimation}
\label{alg:dtde}
\begin{algorithmic}[1]
\Require 
  Streaming batches $\{\boldsymbol{X}^t\}_{t=1}^T$; CLIP model $(\boldsymbol{f}_v, \boldsymbol{f}_t)$; CLIP text prototypes $\{\boldsymbol{w}_k\}_{k=1}^K$; Sketch prediction $\mathrm{P}_{\mathrm{sketch}}$
\Ensure 
  GMM parameters $\{\boldsymbol{\mu}_k^t, \boldsymbol{\Sigma}_k^t, \pi_k^t\}_{k=1}^K$ for each time $t$
\State Initialize means $\boldsymbol{\mu}_k^0$ and covariances $\boldsymbol{\Sigma}_k^0$ using Eq. \eqref{eq:initialize}
\For{each time step $t = 1$ to $T$}
  \State Extract features: $\{\boldsymbol{z}_i^t\}_{i=1}^{N_t} \gets \boldsymbol{f}_v(\boldsymbol{X}^t)$
  \State Compute responsibilities $\gamma_{ik}^t$ via Eq. \eqref{eq:e-step}
  \State Update means $\boldsymbol{\mu}_k^t$ via Eq. \eqref{eq:mu-update}
  \If{$t = 1$}
    \State Determine covariance structure via covariance homogeneity test (see \textbf{Section} \ref{subsection:homogeneity_test})
  \EndIf
  \State Update covariances $\boldsymbol{\Sigma}_k^t$ via Eq. \eqref{eq:cov-homogeneous} or Eq. \eqref{eq:cov-specific}
  \State Regularize covariances via Eq. \eqref{eq:cov-regularize}
  \State Update class priors $\pi_k^t$ via Eq. \eqref{eq:pi-update}
  \State \textbf{Output} $\{\boldsymbol{\mu}_k^t, \boldsymbol{\Sigma}_k^t, \pi_k^t\}_{k=1}^K$
\EndFor
\end{algorithmic}
\end{algorithm}

BayesTTA supports temporally consistent adaptation under evolving test distributions by maintaining a dynamic estimate of the class-conditional generative model described in \textbf{Section} \ref{subsection:overview}. Building on the widely validated assumption that deep features follow class-conditional Gaussian mixture distributions \cite{bishop2006pattern, wang2024a, han2024dota}, it models each class as a multivariate Gaussian with time-dependent parameters:
\begin{equation}
p_t(\boldsymbol{z} \mid Y = k) = \mathcal{N}(\boldsymbol{\mu}_k^t, \boldsymbol{\Sigma}_k^t).
\label{eq:class-conditional}
\end{equation}
Combining these, the marginal feature distribution forms a Gaussian mixture weighted by the class priors $\pi_k^t$:
\begin{equation}
p_t(\boldsymbol{z}) = \sum_{k=1}^K \pi_k^t  \mathcal{N}(\boldsymbol{\mu}_k^t, \boldsymbol{\Sigma}_k^t).
\label{eq:marginal}
\end{equation}
This formulation provides an interpretable and principled foundation for tracking non-stationary distributions.

To capture long-term dynamics without retaining raw data, BayesTTA adopts an incremental Expectation-Maximization (EM) algorithm \cite{cappe2009line}, which propagates sufficient statistics over time. This mechanism enables temporally smoothed, debiased distribution estimation—preserving long-range memory while suppressing short-term noise caused by distribution shifts. At initialization ($t{=}0$), BayesTTA sets each class mean to the corresponding zero-shot CLIP prototype $\boldsymbol{w}_k$, and initializes each covariance matrix as an identity matrix:
\begin{equation}
    \boldsymbol{\mu}_k^{0} = \boldsymbol{w}_k, \quad
    \boldsymbol{\Sigma}_k^{0} = \boldsymbol{I},
    \label{eq:initialize}
\end{equation}
where $\boldsymbol{I}$ is the identity matrix ensuring numerical stability.

At each time step $t$, BayesTTA begins with the E-step by computing the posterior responsibility $\gamma_{ik}^t$, which quantifies the soft assignment probability of sample $i$ to class $k$ under the current generative model:
\begin{equation}
\gamma_{ik}^t = \frac{\pi_k^{t-1} \mathcal{N}(\boldsymbol{z}_i^{t} \mid \boldsymbol{\mu}_k^{t-1}, \boldsymbol{\Sigma}_k^{t-1})}{\sum_{j=1}^K \pi_j^{t-1} \mathcal{N}(\boldsymbol{z}_i^{t} \mid \boldsymbol{\mu}_j^{t-1}, \boldsymbol{\Sigma}_j^{t-1})}.
\label{eq:e-step}
\end{equation}

The M-step then incrementally updates model parameters by accumulating sufficient statistics over time. The soft count for class $k$ is updated as:
\begin{equation}
s_k^{t} = s_k^{t-1} + \sum_{i=1}^{N_t} \gamma_{ik}^t,
\label{eq:soft-count}
\end{equation}
where $ s_k^0 {=} 1 $ is a small initialization constant to ensure numerical stability. This quantity reflects the cumulative confidence mass of class $k$ up to time $t$. The class mean is then updated via weighted averaging:
\begin{equation}
\boldsymbol{\mu}_k^{t} = \frac{s_k^{t-1} \boldsymbol{\mu}_k^{t-1} + \sum_{i=1}^{N_t} \gamma_{ik}^t \boldsymbol{z}_i^t}{s_k^{t}}.
\label{eq:mu-update}
\end{equation}

The covariance update depends on the outcome of the covariance homogeneity test (see \textbf{Section} \ref{subsection:homogeneity_test}). In the homogeneous case, a shared covariance matrix across classes is maintained and updated as:
\begin{equation}
\boldsymbol{\Sigma}^{t} = \frac{s^{t-1} \boldsymbol{\Sigma}^{t-1} + \sum_{i=1}^{N_t} \sum_{k=1}^K \gamma_{ik}^t \Delta \boldsymbol{z}_{ik}^t (\Delta \boldsymbol{z}_{ik}^t)^\top}{s^{t-1} + N_t}, 
\label{eq:cov-homogeneous}
\end{equation}
where $s^{t-1} {=} \sum_{k=1}^K s_k^{t-1}$, $\Delta \boldsymbol{z}_{ik}^t {=} \boldsymbol{z}_i^t-\boldsymbol{\mu}_k^{t}$, and $\boldsymbol{\Sigma}^0$ is initialized as the average covariance across classes at $t{=}0$. Otherwise, heterogeneous covariances are updated as:
\begin{equation}
\boldsymbol{\Sigma}_k^{t} = \frac{s_k^{t-1} \boldsymbol{\Sigma}_k^{t-1} + \sum_{i=1}^{N_t} \gamma_{ik}^t \Delta \boldsymbol{z}_{ik}^t (\Delta \boldsymbol{z}_{ik}^t)^\top}{s_k^{t}}. 
\label{eq:cov-specific}
\end{equation}

A ridge-style regularization is applied to ensure numerical stability:
\begin{equation}
\boldsymbol{\Sigma}_k^{t} \leftarrow (1 - \epsilon)\, \boldsymbol{\Sigma}_k^{t} + \epsilon\, \sigma_{\mathrm{prior}}^2 \boldsymbol{I},
\label{eq:cov-regularize}
\end{equation}
where $\epsilon$ controls the regularization strength and $\sigma_{\mathrm{prior}}^2$ is a fixed heuristic approximating average feature variance (e.g., 0.1). In the homogeneous case, all $\boldsymbol{\Sigma}_k^{t}$ are set equal to $\boldsymbol{\Sigma}^{t}$.

Class priors are updated by normalizing the cumulative soft counts:
\begin{equation}
\pi_k^{t} = \frac{s_k^{t}}{\sum_{j=1}^K s_j^{t}}.
\label{eq:pi-update}
\end{equation}

Although updated incrementally, the accumulated sufficient statistics serve to progressively smooth out local fluctuations. Under gradual temporal drift—typical in CT-TTA scenarios—this mechanism induces a convergent generative trajectory. This behavior aligns with theoretical results from online EM \cite{cappe2009line}, where continual updates support stable adaptation while preserving long-term distributional memory, thereby enhancing robustness and mitigating catastrophic forgetting.

\subsection{Sketch Refinement via Gaussian Discriminant Analysis}
\label{subsection:sketch_refinement}
To enhance CLIP's generalization under temporally evolving distribution shifts, we propose a Bayesian framework grounded in Gaussian discriminant analysis (GDA) \cite{hastie2009elements} to calibrate sketch predictions. This component integrates GDA-based prediction calibration, distribution-aware logit fusion, and self-paced test-time optimization, which together enable adaptive alignment of sketch predictions with the evolving target distribution in a principled manner.

\subsubsection{GDA-based prediction calibration}
Assuming class-conditional Gaussian mixture distributions, GDA imposes an explicit probabilistic structure on the feature space that acts as an inductive bias to mitigate overfitting in the absence of supervision. Beyond modeling class means, GDA captures second-order statistics—specifically, intra-class covariance structures—that reflect how feature dimensions co-vary within each class. This allows the model to account for feature anisotropy and heteroscedasticity, commonly overlooked by linear similarity-based approaches such as $\boldsymbol{w}_k^\top \boldsymbol{z}_i^t$ used in TDA \cite{karmanov2024efficient}, which utilize only first-order information. To accommodate different distributional conditions, BayesTTA adaptively selects between two GDA variants: \textit{linear discriminant analysis (LDA)}, which assumes homogeneous covariance across all classes and favors robust, regularized estimates in low-data regimes; and \textit{quadratic discriminant analysis (QDA)}, which allows heterogeneous class-specific covariances to better capture complex structure when sufficient data are available. This adaptive selection mechanism ensures accurate parameter estimation and forms the foundation for reliable temporal calibration in downstream adaptation.

Based on the covariance homogeneity test (\textbf{Section} \ref{subsection:homogeneity_test}), BayesTTA adaptively selects the appropriate covariance structure for each scenario. Specifically, the QDA \textit{discriminant score} for class $k$ at time $t$ is computed as:
\begin{equation}
\mathcal{D}_k(\boldsymbol{z}_i^t) = \log \pi_k^{t} - \frac{1}{2}(\Delta \boldsymbol{z}_{ik}^t)^\top \left( \boldsymbol{\Sigma}_k^{t} \right)^{-1} \Delta \boldsymbol{z}_{ik}^t - \frac{1}{2} \log \left| \boldsymbol{\Sigma}_k^{t} \right|.
\label{eq:log_discriminant}
\end{equation}
Here, $ \left| \boldsymbol{\Sigma}_k^t \right| $ is the determinant of the covariance matrix, and $\boldsymbol{\mu}_k^{t}$, $\boldsymbol{\Sigma}_k^{t}$ and $\pi_k^{t}$ are estimated online via the incremental EM algorithm (\textbf{Section} \ref{subsection:distribution_estimation}). A detailed derivation of Eq. \eqref{eq:log_discriminant} is provided in \textbf{Appendix} A. 

To ensure numerical stability when computing the inverse of $\boldsymbol{\Sigma}_k^t$, we apply ridge-style shrinkage regularization as in Eq. \eqref{eq:cov-regularize}, shrinking covariance estimates toward scaled identity matrices, and use the Moore–Penrose pseudo-inverse \cite{penrose1955generalized}. This approach enables robust estimation even for ill-conditioned or near-singular covariance matrices. When covariances are tested as homogeneous, Eq. \eqref{eq:log_discriminant} simplifies to the LDA form by setting:
\begin{equation}
\boldsymbol{\Sigma}_k^{t} \triangleq \boldsymbol{\Sigma}^{t}, \quad \text{for} \; \forall \; k \in [1,K].
\end{equation}

\subsubsection{Distribution-Aware Logit Fusion}
To leverage pretrained CLIP's knowledge while improving adaptation to evolving distributions, we propose a distribution-aware logit fusion mechanism that integrates static sketch logits with dynamic GDA calibration. This fusion injects class-conditional structure into the decision process, enhancing robustness and mitigating catastrophic forgetting. Given a test feature $\boldsymbol{z}_i^t$, let $\ell_k^{\mathrm{sketch}}(\boldsymbol{z}_i^t)$ denote the CLIP-derived sketch logits, and $\mathcal{D}_k(\boldsymbol{z}_i^t)$ the GDA-based discriminant score. The adapted logits are computed by:
\begin{equation}
\ell_{k}^{\mathrm{adapt}}(\boldsymbol{z}_i^t) = \ell_{k}^{\mathrm{sketch}}(\boldsymbol{z}_i^t) + \alpha \cdot \mathcal{D}_k(\boldsymbol{z}_i^t),
\label{eq:logits_fusion}
\end{equation}
where fusion weight $\alpha$ balances the influence of semantic priors and distribution-aware calibration. Its effect is analyzed in \textbf{Section} \ref{subsection:ablation-studies}. The final class prediction is:
\begin{equation}
\hat{y}_i^t = \arg\max_{k} \ell_{k}^{\mathrm{adapt}}(\boldsymbol{z}_i^t).
\label{eq:prediction}
\end{equation}

This fusion yields robust pseudo-labels without requiring ground-truth annotations, facilitating reliable adaptation under non-stationary test-time conditions.

\subsubsection{Self-Paced Test-Time Optimization}
We introduce a self-paced learning strategy to dynamically adapt CLIP's image encoders, aiming to progressively refine its visual representations in response to evolving inputs. By treating adapted logits as fixed pseudo-labels to supervise the sketch logits—with gradient updates restricted solely to the sketch head—the model iteratively enhances its predictions through self-calibration. This mechanism obviates the need for external annotations or manual sample selection. The refinement objective minimizes the soft cross-entropy between normalized adapted and sketch logits:
\begin{equation}
\mathcal{L}_{\text{refine}} = 
- \sum_{k=1}^{K} 
\mathrm{softmax}(\boldsymbol{\ell}^{\mathrm{adapt}})_k \cdot 
\log \mathrm{softmax}(\boldsymbol{\ell}^{\mathrm{sketch}})_k,
\label{eq:refine_soft_ce}
\end{equation}
where $(\boldsymbol{\ell}^{\mathrm{adapt}})_k$ denotes the $k$-th element of the adapted logits vector, consistent with $\ell_k^{\mathrm{adapt}}(\boldsymbol{z}_i^t)$.

For efficiency, only the normalization layers—LayerNorm \cite{ba2016layer} in ViT and BatchNorm \cite{ioffe2015batch} in ResNet—are updated, while the rest of the image encoder parameters remain frozen. This selective fine-tuning modifies no more than \textbf{0.05\%} of the total model parameters (e.g., for ViT-B/16), reducing computational cost and the risk of overfitting. To further stabilize adaptation, we maintain exponential moving averages (EMA) \cite{polyak1992acceleration} of the model parameters across time steps:
\begin{equation}
\boldsymbol{\theta}_{\mathrm{ema}}^{t} = \beta \cdot \boldsymbol{\theta}_{\mathrm{ema}}^{t-1} + (1 - \beta) \cdot \boldsymbol{\theta}^{t},
\label{eq:ema}
\end{equation}
where $\beta$ is the EMA decay rate. This technique smooths parameter updates and enhances adaptation stability over time.

Together, selective fine-tuning and EMA smoothing enable efficient and stable representation alignment with evolving inputs.

\section{Experiments}

\subsection{Experimental Settings}
\label{subsection:experiments-setup}

\textbf{Datasets.} 
To evaluate the effectiveness of BayesTTA under temporally evolving distribution shifts, we construct a CT-TTA benchmark comprising four datasets: two real-world (fMoW and Yearbook \cite{yao2022wild}) and two synthetic datasets (RMNIST and CIFAR10-Light). These datasets cover diverse visual domains and exhibit various types of gradual domain shifts. A summary is provided in \textbf{Appendix} B, with detailed descriptions below:
\begin{itemize}
\item The \textbf{Functional Map of the World (fMoW)} dataset \cite{yao2022wild} contains satellite images collected from 2002 to 2017 across over 200 countries, capturing visual changes due to environmental and human-driven factors. It defines a 62-class classification task over 16 yearly domains, modeling long-term distribution shifts in remote sensing.
\item The \textbf{Yearbook} dataset \cite{yao2022wild}, adapted from the Portraits dataset \cite{ginosar2015century}, consists of grayscale portrait images spanning 84 consecutive years (1930–2013) from 128 American high schools. It reflects changes in appearance and fashion, with each year as a distinct domain, forming 84 fine-grained temporally evolving distributions.
\item The \textbf{RMNIST} dataset, a rotated variant of MNIST \cite{deng2012mnist}, comprises nine domains generated by rotating digit images from $0^\circ$ to $80^\circ$ in $10^\circ$ increments. This setup simulates smooth evolving distribution shifts over time through systematic geometric transformations.
\item The \textbf{CIFAR10-Light} dataset, proposed in this work, is derived from CIFAR-10 \cite{krizhevsky2009learning} via amplitude-phase recomposition in the Fourier domain \cite{brigham1988fast}. Seven amplitude scaling factors (0.25 to 1.75) simulate lighting conditions ranging from low-light to overexposed, reflecting gradual illumination changes typical of autonomous driving from day to night. This creates seven smoothly evolving domains reflecting varying visual environments. Representative samples are provided in \textbf{Appendix} B.
\end{itemize}

To evaluate the generalization ability of BayesTTA beyond CT-TTA, we further conduct experiments under the TTA setting. Following prior work \cite{karmanov2024efficient}, we adopt the cross-domain benchmark consisting of 10 diverse image classification datasets, each representing a distinct domain: Aircraft \cite{maji2013fine}, Caltech101 \cite{fei2004learning}, StanfordCars \cite{krause20133d}, DTD \cite{cimpoi2014describing}, EuroSAT \cite{helber2019eurosat}, Flower102 \cite{nilsback2008automated}, Food101 \cite{bossard2014food}, OxfordPets \cite{parkhi2012cats}, SUN397 \cite{xiao2010sun}, and UCF101 \cite{soomro2012ucf101}. 

For CT-TTA, accuracy is computed as a weighted average over all temporal domains, with weights proportional to domain sample sizes, thereby reflecting the sequential test stream distribution. In contrast, for standard TTA, we report the unweighted average accuracy across ten distinct domains, consistent with prior work that treats each domain independently without temporal ordering.

\textbf{Implementation details.} Our experiments are built upon the pre-trained CLIP architecture \cite{radford2021learning}, which comprises a dual-encoder design: an image encoder (ResNet-50 \cite{he2016deep} or ViT-B/16 \cite{dosovitskiy2020image}) and a Transformer-based text encoder \cite{vaswani2017attention}. The batch size is set to 128, following WATT \cite{osowiechi2024watt} and CoTTA \cite{wang2022continual}, to ensure a fair comparison. This setting balances computational efficiency and statistical reliability in covariance estimation, as confirmed by the ablation study (\textbf{Section} \ref{subsection:ablation-studies}). To avoid the computational overhead of prompt tuning, we adopt hand-crafted prompts following \cite{radford2021learning}, using the same template as TDA \cite{karmanov2024efficient} for consistency. All models are evaluated using top-1 accuracy (\%), and all experiments are conducted on a single NVIDIA RTX 3090 GPU.

\textbf{Baselines.} We compare our proposed method with the following SOTA methods. TENT \cite{wang2021tent}, the first method to employ test-time entropy minimization via online parameter updates. CoTTA \cite{wang2022continual}, the first CTTA approach via weight-averaged and augmentation-averaged predictions. RoTTA \cite{yuan2023robust}, robustness for time-correlated sampling. TPT \cite{shu2022test}, a test-time prompt tuning method that adapts prompts by minimizing self-entropy over augmented images. Adaprompt \cite{wang2024towards}, prompt ensembling and a confidence-aware buffer for addressing data and model bias. TDA \cite{karmanov2024efficient}, a training-free adapter-based method that builds positive and negative caches during inference. WATT \cite{osowiechi2024watt} utilizes weight averaging over multiple text templates as a self-supervised optimization objective. DPE \cite{zhang2024dual}, dual-modal prototype evolution for incremental feature-space alignment. DynaPrompt \cite{xiao2025dynaprompt}, dynamic test-time prompt tuning via beneficial information selection. To align with prior work \cite{karmanov2024efficient}, we also report ensemble results using 80 handcrafted prompts \cite{radford2021learning}. Additionally, we compare our method with DOTA \cite{han2024dota}, HisTPT \cite{zhang2024historical}, and BCA \cite{zhou2025bayesian} on the cross-domain benchmark under the standard TTA setting. Since their official code is not publicly available, we report the results as presented in their original publications.

\subsection{Comparisons with State-of-the-art}
\label{subsection:comparisons-with-state-of-the-art}

\begin{table*}
\centering
\caption{Performance comparisons on the fMoW dataset under CT-TTA paradigm. The best results are highlighted in bold.}
\label{tab:fmow-datasets}
\tabcolsep=0.12cm
\begin{tabular}{lccccccccccccccccc}
\toprule[1pt]
Time & \multicolumn{16}{c}{t}{$\xrightarrow{\hspace{13cm}}$} \\
\midrule
\diagbox[width=2.8cm]{Method}{Year} & 2002 & 2003 & 2004 & 2005 & 2006 & 2007 & 2008 & 2009 & 2010 & 2011 & 2012 & 2013 & 2014 & 2015 & 2016 & 2017 & Average\\
\midrule[1pt]
CLIP-ResNet-50 \cite{radford2021learning} & 6.25 & 6.81 & 8.02 & 6.62 & 8.81 & 8.60 & 7.07 & 9.06 & 13.13 & 14.03 & 11.58 & 12.36 & 14.05 & 13.17 & 14.64 & 14.59 & 12.32\\
Ensemble & 6.52 & 7.32 & 8.52 & 7.40 & 9.60 & 8.77 & 7.27 & 9.15 & 13.50 & 14.30 & 11.57 & 12.83 & 14.45 & 13.35 & 14.83 & 15.45 & 12.59 \\
TENT-reset \cite{wang2021tent} & 6.31 & 6.85 & 7.92 & 7.26 & 9.10 & 8.65 & 6.88 & 9.22 & 13.49 & 14.49 & 11.67 & 13.14 & 14.24 & 13.21 & 14.62 & 14.03 & 12.50 \\
TENT-continual \cite{wang2021tent} & 3.36 & 2.35 & 3.61 & 3.46 & 5.08 & 4.03 & 2.46 & 2.42 & 3.94 & 3.55 & 3.20 & 2.94 & 2.55 & 3.43 & 3.70 & 5.63 & 3.51\\
CoTTA \cite{wang2022continual} & 3.99 & 3.64 & 1.95 & 1.20 & 1.26 & 1.54 & 1.33 & 1.97 & 1.31 & 0.95 & 1.11 & 0.88 & 0.94 & 1.07 & 1.05 & 0.39 & 1.22\\
RoTTA \cite{yuan2023robust} & 6.41 & 6.93 & 8.12 & 6.10 & 8.65 & 5.51 & 2.89 & 2.46 & 3.63 & 2.10 & 2.31 & 2.57 & 2.28 & 3.37 & 3.73 & 5.55 & 3.37 \\
TPT \cite{shu2022test} & 6.31 & 7.32 & 8.32 & 6.59 & 8.71 & 8.54 & 7.07 & 9.20 & 13.39 & 14.50 & 11.53 & 13.14 & 14.29 & 13.67 & 14.62 & 14.08 & 12.49 \\
Adaprompt \cite{wang2024towards} & 6.31 & 5.48 & 6.62 & 6.28 & 6.88 & 7.41 & 6.25 & 8.01 & 12.53 & 13.95 & 12.22 & 13.53 & 13.97 & 14.27 & 15.24 & 14.02 & 12.29 \\
TDA \cite{karmanov2024efficient} & 7.83 & 8.06 & 8.22 & 8.22 & 10.50 & 10.67 & 9.57 & 11.97 & 14.57 & 14.90 & 12.82 & 14.42 & 13.99 & 14.30 & 15.97 & 14.65 & 13.58 \\
WATT \cite{osowiechi2024watt} & 6.99 & 7.98 & 7.57 & 7.37 & 10.00 & 9.19 & 7.50 & 9.74 & 11.88 & 13.26 & 12.73 & 13.77 & 14.10 & 14.08 & 14.36 & 11.21 & 12.27 \\
DPE \cite{zhang2024dual} & 6.41 & 7.28 & 7.92 & 6.70 & 8.51 & 8.18 & 7.07 & 9.86 & 13.49 & 14.43 & 12.11 & 13.12 & 13.91 & 13.16 & 14.81 & 12.67 & 12.54 \\
DynaPrompt \cite{xiao2025dynaprompt} & 5.78 & 6.65 & 7.87 & 6.38 & 8.45 & 7.82 & 6.41 & 8.71 & 12.74 & 13.80 & 10.91 & 12.65 & 13.57 & 12.93 & 14.00 & 13.11 & 11.85 \\
\rowcolor{grayblue}
\textbf{BayesTTA(Ours)} & \textbf{8.57} & \textbf{10.53} & \textbf{10.93} & \textbf{10.65} & \textbf{12.13} & \textbf{11.26} & \textbf{10.98} & \textbf{12.17} & \textbf{15.76} & \textbf{17.11} & \textbf{14.82} & \textbf{16.13} & \textbf{16.33} & \textbf{16.19} & \textbf{18.17} & \textbf{16.77} & \textbf{15.39} \\
\midrule \midrule
CLIP-ViT-B/16 \cite{radford2021learning} & 14.77 & 13.11 & 13.13 & 12.48 & 16.20 & 13.93 & 14.62 & 16.13 & 19.67 & 22.40 & 20.37 & 20.16 & 22.74 & 21.45 & 23.01 & 22.25 & 20.17 \\
Ensemble & 13.35 & 13.11 & 13.13 & 12.31 & 15.19 & 14.17 & 14.77 & 16.10 & 20.36 & 23.09 & 21.05 & 21.30 & 23.64 & 22.09 & 23.45 & 22.69 & 20.67 \\
TENT-reset \cite{wang2021tent} & 13.50 & 12.41 & 12.73 & 11.99 & 15.47 & 13.28 & 14.50 & 16.06 & 20.38 & 23.09 & 21.07 & 21.04 & 23.42 & 21.91 & 23.19 & 22.02 & 20.56 \\
TENT-continual \cite{wang2021tent} & 10.98 & 4.03 & 3.71 & 3.46 & 5.08 & 4.03 & 2.46 & 2.42 & 3.94 & 3.55 & 3.20 & 2.94 & 2.55 & 3.43 & 3.70 & 5.63 & 3.64 \\
CoTTA \cite{wang2022continual} & 4.99 & 6.03 & 5.11 & 4.06 & 3.57 & 2.49 & 1.45 & 1.27 & 1.96 & 2.67 & 2.77 & 1.79 & 1.42 & 1.14 & 1.75 & 1.85 & 2.33 \\
RoTTA \cite{yuan2023robust} & 13.45 & 12.37 & 13.23 & 11.53 & 14.80 & 13.34 & 13.76 & 15.81 & 20.21 & 22.78 & 21.27 & 21.30 & 24.16 & 22.56 & 23.36 & \textbf{23.08} & 20.62 \\
TPT \cite{shu2022test} & 13.19 & 13.23 & 13.03 & 12.55 & 14.66 & 13.87 & 14.97 & 17.02 & 21.14 & 23.42 & 21.08 & 21.82 & 23.79 & 22.82 & 23.90 & 21.39 & 20.96 \\
Adaprompt \cite{wang2024towards} & 12.72 & 10.65 & 9.87 & 11.32 & 13.17 & 12.86 & 13.64 & 14.43 & 19.52 & 22.08 & 20.66 & 21.45 & 22.27 & 22.23 & 22.51 & 19.94 & 19.71 \\
TDA \cite{karmanov2024efficient} & 16.61 & 13.78 & 12.33 & 14.00 & 17.97 & 15.29 & 17.04 & 18.51 & 20.08 & 22.63 & 21.31 & 21.90 & 21.08 & 20.56 & 21.91 & 20.39 & 20.47 \\
WATT \cite{osowiechi2024watt} & 15.24 & 14.13 & 13.38 & 14.84 & 17.78 & 16.78 & 17.16 & 18.26 & 21.32 & 24.12 & 22.67 & 23.27 & 24.03 & 23.60 & 24.41 & 20.96 & 21.89 \\
DPE \cite{zhang2024dual} & 10.93 & 11.39 & 12.63 & 11.18 & 14.94 & 13.87 & 14.19 & 15.44 & 18.91 & 21.22 & 19.85 & 19.01 & 20.16 & 19.64 & 21.30 & 17.78 & 18.90 \\
DynaPrompt \cite{xiao2025dynaprompt} & 12.30 & 12.68 & 11.58 & 12.13 & 13.68 & 12.92 & 14.34 & 15.78 & 20.43 & 22.56 & 20.32 & 21.30 & 23.42 & 21.16 & 22.41 & 21.27 & 20.06 \\
\rowcolor{grayblue}
\textbf{BayesTTA(Ours)} & \textbf{16.82} & \textbf{14.83} & \textbf{15.04} & \textbf{17.56} & \textbf{19.80} & \textbf{18.91} & \textbf{18.95} & \textbf{20.11} & \textbf{22.91} & \textbf{25.61} & \textbf{24.54} & \textbf{26.13} & \textbf{25.50} & \textbf{25.61} & \textbf{26.16} & 22.49 & \textbf{23.63} \\
\bottomrule[1pt]
\end{tabular}
\end{table*}

\begin{table}
\centering
\caption{Performance comparisons on the Yearbook dataset under CT-TTA paradigm. The best results are highlighted in bold.}
\label{tab:yearbook-dataset}
\begin{tabular}{lcc}
\toprule[1pt]
Time & \multicolumn{1}{c}{t}{$\xrightarrow{\hspace{2cm}}$} \\
\midrule
Method & Average {\tiny (Year 1930-2013)} & Gains \\
\midrule[1pt]
CLIP-ResNet-50 \cite{radford2021learning} & 90.64 & - \\
Ensemble & 90.64 & 0.00 \\
TENT-reset \cite{wang2021tent} & 90.67 & +0.03\\
TENT-continual \cite{wang2021tent} & 53.33 & -37.31\\
CoTTA \cite{wang2022continual} & 68.83 & -21.81\\
RoTTA \cite{yuan2023robust} & 71.32 & -19.32\\
TPT \cite{shu2022test} & 89.57 & -1.07\\
Adaprompt \cite{wang2024towards} & 90.63 & -0.01\\
TDA \cite{karmanov2024efficient} & 90.40 & -0.24\\
WATT \cite{osowiechi2024watt} & 90.81 & +0.17\\
DPE \cite{zhang2024dual} & 81.58 & -9.06\\
DynaPrompt \cite{xiao2025dynaprompt} & 87.22 & -3.42\\
\rowcolor{grayblue}
\textbf{BayesTTA(Ours)} & \textbf{91.73} & \textbf{+1.09}\\
\midrule \midrule
CLIP-ViT-B/16 \cite{radford2021learning} & 95.76 & -\\
Ensemble & 95.71 & -0.05\\
TENT-reset \cite{wang2021tent} & 95.67 & -0.09\\
TENT-continual \cite{wang2021tent} & 95.83 & +0.07\\
CoTTA \cite{wang2022continual} & 88.80 & -6.96\\
RoTTA \cite{yuan2023robust} & 95.94 & +0.18\\
TPT \cite{shu2022test} & 95.54 & -0.22\\
Adaprompt \cite{wang2024towards} & 95.84 & +0.08\\
TDA \cite{karmanov2024efficient} & 95.69 & -0.07\\
WATT \cite{osowiechi2024watt} & 96.08 & +0.32\\
DPE \cite{zhang2024dual} & 90.06 & -5.70\\
DynaPrompt \cite{xiao2025dynaprompt} & 92.91 & -2.85\\
\rowcolor{grayblue}
\textbf{BayesTTA(Ours)} & \textbf{96.29} & \textbf{+0.53}\\
\bottomrule[1pt]
\end{tabular}
\end{table}

\begin{table*}
\centering
\caption{Performance comparisons on RMNIST dataset under CT-TTA paradigm. The best results are highlighted in bold.}
\label{tab:rmnist-datasets}
\begin{tabular}{lcccccccccc}
\toprule[1pt]
Time & \multicolumn{9}{c}{t}{$\xrightarrow{\hspace{8.5cm}}$} \\
\midrule
\diagbox[width=2.6cm]{Method}{Angle} & $0^\circ$ & $10^\circ$ & $20^\circ$ & $30^\circ$ & $40^\circ$ & $50^\circ$ & $60^\circ$ & $70^\circ$ & $80^\circ$ & Average\\
\midrule[1pt]
CLIP-ResNet-50 \cite{radford2021learning} & 49.51 & 47.87 & 40.91 & 31.31 & 22.72 & 16.83 & 13.42 & 13.96 & 13.40 & 27.77 \\
Ensemble & 49.07 & 47.85 & 43.70 & 37.62 & 31.47 & 27.11 & 26.42 & 26.23 & 23.13& 34.73\\
TENT-reset \cite{wang2021tent} & 54.08 & 54.83 & 50.95 & 43.39 & 35.64 & 29.79 & 26.90 & 25.74 & 24.44 & 38.42 \\
TENT-continual \cite{wang2021tent} & 14.45  & 9.33  & 9.76  & 10.05  & 10.08  & 9.68  & 10.04  & 9.18  & 9.79 & 10.26 \\
CoTTA \cite{wang2022continual} & 15.35 & 8.32 & 8.36 & 8.63 & 8.60 & 9.84 & 9.94 & 11.87 & 11.98 & 10.32 \\
RoTTA \cite{yuan2023robust} & 54.32 & 44.55 & 17.02 & 10.23 & 10.56 & 10.90 & 10.52 & 9.48 & 9.68 & 19.70 \\
TPT \cite{shu2022test} & 37.67 & 34.01 & 27.59 & 23.10 & 18.63 & 16.79 & 15.67 & 15.75 & 15.17 & 22.71 \\
Adaprompt \cite{wang2024towards} & 42.63 & 40.63 & 31.99 & 23.30 & 20.96 & 17.45 & 16.12 & 14.36 & 12.60 & 24.45 \\
TDA \cite{karmanov2024efficient} & 57.28 & 52.40 & 46.03 & 40.38 & 34.66 & 28.25 & 24.97 & 22.64 & 21.62 & 36.47 \\
WATT \cite{osowiechi2024watt} &  65.30 & 66.77 & 63.58 & 56.92 & 41.23 & 30.57 & 21.00 & 16.46 & 15.08 & 41.88 \\
DPE \cite{zhang2024dual} & 16.48 & 14.36 & 15.11 & 11.07 & 13.89 & 12.51 & 16.75 & 18.93 & 19.49 & 15.40 \\
DynaPrompt \cite{xiao2025dynaprompt}  & 40.14 & 36.23 & 29.83 & 25.14 & 19.67 & 17.19 & 15.44 & 16.20 & 12.23 & 23.56 \\
\rowcolor{grayblue}
\textbf{BayesTTA(Ours)} & \textbf{70.21} & \textbf{81.15} & \textbf{81.95} & \textbf{81.07} & \textbf{77.54} & \textbf{76.09} & \textbf{71.73} & \textbf{69.47} & \textbf{66.45} & \textbf{75.07} \\
\midrule \midrule
CLIP-ViT-B/16 \cite{radford2021learning} & 45.11 & 44.24 & 37.58 & 28.07 & 21.38 & 18.54 & 17.09 & 15.79 & 13.90 & 26.86 \\
Ensemble & 56.75 & 54.32 & 45.50 & 35.21 & 26.56 & 22.59 & 18.41 & 15.07 & 11.75& 31.80 \\
TENT-reset \cite{wang2021tent} & 56.96 & 54.73 & 45.23 & 34.28 & 27.09 & 23.70 & 20.91 & 19.07 & 16.36 & 33.15 \\
TENT-continual \cite{wang2021tent} & 64.19  & 46.84  & 20.21  & 10.05  & 10.08  & 9.68 & 10.04  & 9.18  & 9.79 & 21.12 \\
CoTTA \cite{wang2022continual} & 30.82 & 27.37 & 23.17 & 19.50 & 17.04 & 13.96 & 13.33 & 11.74 & 11.60 & 18.73 \\
RoTTA \cite{yuan2023robust} & 57.37 & 57.74 & 49.56 & 38.04 & 27.41 & 23.23 & 22.28 & 21.40 & 19.90 & 35.22 \\
TPT \cite{shu2022test} & 53.14 & 51.61 & 41.61 & 31.95 & 24.04 & 20.40 & 18.09 & 15.39 & 13.62 & 29.98 \\
Adaprompt \cite{wang2024towards} & 62.74 & 59.76 & 51.21 & 39.32 & 29.57 & 23.72 & 17.31 & 14.05 & 11.52 & 34.36 \\
TDA \cite{karmanov2024efficient} & 55.18 & 60.31 & 53.39 & 43.78 & 35.95 & 30.44 & 26.20 & 23.45 & 21.06 & 38.86 \\
WATT \cite{osowiechi2024watt} & \textbf{78.73} & \textbf{76.73} & 74.18 & 73.36 & 69.54 & 52.98 & 34.48 & 26.64 & 20.98 & 56.40 \\
DPE \cite{zhang2024dual} & 11.25 & 16.29 & 46.90 & 32.42 & 23.25 & 19.44 & 19.40 & 18.77 & 17.56 & 22.81 \\
DynaPrompt \cite{xiao2025dynaprompt} & 50.23 & 49.18 & 39.42 & 29.61 & 23.89 & 19.18 & 17.41 & 14.70 & 13.01& 28.51 \\
\rowcolor{grayblue}
\textbf{BayesTTA(Ours)} & 71.24 & 75.38 & \textbf{74.24} & \textbf{76.30} & \textbf{76.92} & \textbf{76.33} & \textbf{71.55} & \textbf{63.84} & \textbf{50.20} & \textbf{70.67} \\
\bottomrule[1pt]
\end{tabular}
\end{table*}

\begin{table}
\centering
\caption{Performance comparisons on the CIFAR10-Light dataset under CT-TTA paradigm. The best results are highlighted in bold.}
\label{tab:cifar10-light-datasets}
\tabcolsep=0.06cm
\resizebox{0.5\textwidth}{!}{
\begin{tabular}{lcccccccc}
\toprule[1pt]
Time & \multicolumn{7}{c}{t}{$\xrightarrow{\hspace{4.5cm}}$} \\
\midrule
\diagbox[width=2.6cm]{Method}{Light Intensity} & 0.25 & 0.50 & 0.75 & 1.00 & 1.25 & 1.50 & 1.75 & Average\\
\midrule[1pt]
CLIP-ResNet-50 \cite{radford2021learning} & 52.53 & 65.67 & 69.46 & 68.83 & 63.56 & 55.13 & 47.33 & 60.35 \\
Ensemble & 54.33 & 68.04 & 72.11 & 72.45 & 68.18 & 60.77 & 53.49 & 64.20 \\
TENT-reset \cite{wang2021tent} & 53.00 & 67.71 & 72.08 & 72.43 & 67.92 & 59.54 & 52.09 & 63.54 \\
TENT-continual \cite{wang2021tent} & 14.74 & 10.00 & 10.00 & 10.00 & 10.00 & 10.00 & 10.00 & 10.68 \\
CoTTA \cite{wang2022continual} & 10.00 & 10.05 & 12.46 & 20.29 & 20.10 & 16.45 & 14.37 & 14.82 \\
RoTTA \cite{yuan2023robust} & 38.16 & 18.83 & 10.28 & 10.00 & 10.00 & 10.00 & 10.00 & 15.32 \\
TPT \cite{shu2022test} & 58.04 & 69.77 & 72.28 & 72.78 & 68.69 & 61.81 & 54.32 & 65.38 \\
Adaprompt \cite{wang2024towards} & 57.87 & 70.96 & 74.38 & 76.09 & 72.22 & 65.94 & 58.40 & 67.98 \\
TDA \cite{karmanov2024efficient} & 56.60 & 72.27 & 74.92 & 74.51 & 71.32 & 64.78 & 58.41 & 67.54 \\
WATT \cite{osowiechi2024watt} & \textbf{75.74} & 79.65 & 80.21 & 80.58 & 78.94 & 75.37 & 71.98 & 77.50 \\
DPE \cite{zhang2024dual} & 35.64 & 69.58 & 60.17 & 69.38 & 57.31 & 58.65 & 51.10 & 57.40 \\
DynaPrompt \cite{xiao2025dynaprompt} & 51.46 & 67.61 & 70.99 & 55.83 & 66.95 & 58.78 & 49.72 & 60.19 \\
\rowcolor{grayblue}
\textbf{BayesTTA(Ours)} & 73.41 & \textbf{83.04} & \textbf{85.47} & \textbf{85.58} & \textbf{85.16} & \textbf{81.63} & \textbf{76.84} & \textbf{81.59} \\
\midrule \midrule
CLIP-ViT-B/16 \cite{radford2021learning} & 83.92 & 88.97 & 89.81 & 89.21 & 87.42 & 83.33 & 77.25 & 85.70 \\
Ensemble & 85.25 & 89.94 & 90.98 & 90.79 & 89.24 & 85.33 & 79.79 & 87.33 \\
TENT-reset \cite{wang2021tent} & 85.32 & 90.25 & 91.15 & 91.04 & 89.50 & 85.53 & 79.75 & 87.51\\
TENT-continual \cite{wang2021tent} & 86.32 & 91.17 &	92.16 &	92.45 &	91.69 &	89.03 &	84.78 &	89.66  \\
CoTTA \cite{wang2022continual} & 10.00 & 11.53 & 20.97 & 31.61 & 38.83 & 39.48 & 35.87 & 26.90 \\
RoTTA \cite{yuan2023robust} & 84.89 & 90.08 & 91.53 & 92.12 & 91.69 & 89.70 & 85.87 & 89.41 \\
TPT \cite{shu2022test} & 85.98 & 89.44 & 90.17 & 89.76 & 88.24 & 84.16 & 78.82 & 86.65 \\
Adaprompt \cite{wang2024towards} & 85.69 & 90.36 & 91.36 & 91.44 & 89.93 & 86.05 & 79.92 & 87.82 \\
TDA \cite{karmanov2024efficient} & 86.27 & 90.74 & 91.83 & 91.58 & 90.10 & 86.69 & 81.58 & 88.40 \\
WATT \cite{osowiechi2024watt}   & 89.94 & 91.69 & 92.17 & 91.90 & 90.86 & 88.85 & 85.48 & 90.13 \\
DPE \cite{zhang2024dual} & 86.44 & 90.92 & 91.33 & 91.34 & 87.66 & 83.58 & 78.45 & 87.10 \\
DynaPrompt \cite{xiao2025dynaprompt} & 85.33 & 83.38 & 88.14 & 77.44 & 87.40 & 78.72 & 73.00 & 81.92 \\
\rowcolor{grayblue}
\textbf{BayesTTA(Ours)} & \textbf{91.03} & \textbf{94.74} & \textbf{95.28} & \textbf{95.56} & \textbf{95.08} & \textbf{93.43} & \textbf{90.67} & \textbf{93.68} \\
\bottomrule[1pt]
\end{tabular}}
\end{table}

\textbf{Results on the fMoW dataset.} 
Table \ref{tab:fmow-datasets} summarizes results on fMoW, a challenging dataset with a large domain gap between natural and remote sensing imagery that severely limits CLIP's zero-shot accuracy (12.32\% for ResNet-50, 20.17\% for ViT-B/16). Conventional CTTA methods such as TENT-continual and CoTTA degrade sharply or collapse over time. In contrast, BayesTTA consistently improves accuracy to 15.39\% and 23.63\%, outperforming the strongest baselines (TDA, WATT) by +1.81\% and +1.74\%, respectively. This validates BayesTTA's effectiveness in debiased temporal distribution estimation and calibrated prediction for robust adaptation in evolving Earth observation domains.

\textbf{Results on the Yearbook dataset.} 
Table \ref{tab:yearbook-dataset} reports average accuracy across 84 yearly domains spanning 1930–2013. BayesTTA achieves the highest accuracy on both backbones, consistently surpassing all baselines. While WATT offers moderate improvements, continual adaptation methods like TENT-continual and CoTTA suffer severe performance drops, highlighting their vulnerability to gradual temporal shifts. BayesTTA's stable gains over decades demonstrate superior robustness and long-term generalization under continuous domain evolution.

\textbf{Results on the RMNIST dataset.} 
CLIP's zero-shot performance on digit datasets like MNIST is poor due to domain mismatch and lack of texture cues \cite{radford2021learning}, further challenged by the systematic rotation drift in RMNIST. As shown in Table \ref{tab:rmnist-datasets}, input distributions evolve with rotations from $0^\circ$ to $80^\circ$. Using ResNet-50, zero-shot CLIP suffers substantial degradation, with accuracy dropping from 49.51\% at $0^\circ$ to 13.40\% at $80^\circ$, revealing its vulnerability to continuous distribution shifts. In contrast, BayesTTA achieves a substantially higher average accuracy of 75.07\%, maintaining 66.45\% accuracy at $80^\circ$, outperforming CLIP and WATT by +53.05\% and +33.19\%, respectively. The highest accuracy occurs near $20^\circ$–$30^\circ$, reflecting BayesTTA's strong adaptation at the domain transition boundary. Baselines like CoTTA, RoTTA, and TENT-continual show unstable or collapsed adaptation. Similar trends on ViT-B/16 further validate BayesTTA's ability to model and adapt to non-stationary dynamics via hypothesis-guided Gaussian discriminant analysis.

\textbf{Results on the CIFAR10-Light dataset.}
Table \ref{tab:cifar10-light-datasets} reports results under progressive illumination shifts, simulating real-world scenarios such as day-to-night transitions in autonomous driving. Zero-shot CLIP exhibits pronounced sensitivity to illumination shifts. For instance, CLIP-ResNet-50 drops from 68.83\% under standard lighting (1.00) to 52.53\% and 47.33\% at extreme intensities (0.25 and 1.75), reflecting poor robustness under long-term illumination shifts. In contrast, BayesTTA achieves superior performance, with an average accuracy of 81.59\% on ResNet-50 and 93.68\% on ViT-B/16, exceeding the strongest baseline (WATT) by +4.09\% and +3.55\%. Notably, BayesTTA maintains over 85\% accuracy under moderate lighting, while CTTA methods like DynaPrompt degrade sharply. These results confirm that GDA-based prediction calibration enables temporally consistent modeling of evolving distributions, supporting robust long-term adaptation.

\begin{table*}
\centering
\caption{Performance comparison on cross-domain benchmark under TTA. The best results are highlighted in bold.}
\label{tab:cross-datasets}
\tabcolsep=0.15cm
\begin{tabular}{lccccccccccc}
\toprule[1pt]
Method & Aircraft & Caltech & Cars & DTD & EuroSAT & Flower & Food101 & Pets & SUN397 & UCF101 & Average\\
\midrule
CLIP-ResNet-50 \cite{radford2021learning} & 15.66 & 85.88 & 55.70 & 40.37 & 23.69 & 61.75 & 73.97 & 83.57 & 58.80 & 58.84 & 55.82\\
Ensemble & 16.11 & 87.26 & 55.89 & 40.37 & 25.79 & 62.77 & 74.82 & 82.97 & 60.85 & 59.48 & 56.63\\
CoOp \cite{zhou2022learning} & 15.12 & 86.53 & 55.32 & 37.29 & 26.20 & 61.55 & 75.59 & 87.00 & 58.15 & 59.05 & 56.18 \\
TENT-reset \cite{wang2021tent} & 16.56 & 85.68 & 56.56 & 40.66 & 26.86 & 63.34 & 72.23 & 82.23 & 59.25 & 54.19 & 55.76 \\
TENT-continual \cite{wang2021tent} & 16.62 & 86.21 & 54.57 & 41.31 & 24.37 & 63.62 & 72.99 & 82.15 & 59.21 & 55.09 & 55.61 \\
CoTTA \cite{wang2022continual} & 6.78 & 70.43 & 19.10 & 28.37 & 15.42 & 32.76 & 14.10 & 51.21 & 29.80 & 28.76 & 29.67 \\
RoTTA \cite{yuan2023robust} & 16.56 & 85.52 & 56.50 & 40.72 & 27.05 & 63.34 & 65.34 & 82.12 & 58.51 & 54.08 & 54.97\\
TPT \cite{shu2022test} & 17.58 & 87.02 & 58.46 & 40.84 & 28.33 & 62.69 & 74.88 & 84.49 & 61.46 & 60.82 & 57.66\\
Adaprompt \cite{wang2024towards} & 15.42 & 88.48 & 54.84 & 40.54 & 30.90 & 65.73 & 75.68 & 83.81 & - & 58.58 & - \\
TDA \cite{karmanov2024efficient} & 17.61 & 89.70 & 57.78 & 43.74 & 42.11 & 68.74 & 77.75 & 86.18 & 62.53 & 64.18 & 61.03\\
WATT \cite{osowiechi2024watt} & 17.94 & 85.52 & 56.59 & 41.90  & 54.28 & 63.38 & 72.53 & 83.54 & 59.21 & 57.26 & 59.22 \\ 
DPE \cite{zhang2024dual} & 19.80 & \textbf{90.83} & 59.26 & \textbf{50.18} & 41.67 & 67.60 & 77.83 & 85.97 & 64.23 & 61.98 & 61.93\\
DynaPrompt \cite{xiao2025dynaprompt} & 16.44 & 88.07 & - & 41.19 & 23.41 & 61.84 & 75.73 & 83.07 & - & 59.87 & -\\
DOTA \cite{han2024dota}	& 18.06 & 88.84 & 58.72 & 45.80 & 47.15 & 68.53 & 78.61 & 87.33 & 63.89 & 65.08 & 62.20 \\
HisTPT \cite{zhang2024historical} &	18.10 &	87.20 &	\textbf{61.30} & 41.30 & 42.50 &	67.60 &	\textbf{81.30} & 84.90 & 63.50 &	64.10 & 61.20 \\
BCA \cite{zhou2025bayesian} & \textbf{19.89} & 89.70 & 58.13 & 48.58 & 42.12 & 66.30 & 77.19 & 85.58 & 63.38 & 63.51 & 61.44 \\
\rowcolor{grayblue}
\textbf{BayesTTA(Ours)} & 17.85 & 89.41 & 61.07 & 46.45 & \textbf{70.00} & \textbf{69.22} & 78.93 & \textbf{88.53} & \textbf{64.34} & \textbf{65.24} & \textbf{65.11} \\
\midrule \midrule
CLIP-ViT-B/16 \cite{radford2021learning} & 23.67 & 93.35 & 65.48 & 44.27 & 42.01 & 67.44 & 83.65 & 88.25 & 62.59 & 65.13 & 63.58\\
Ensemble & 23.22 & 93.55 & 66.11 & 45.04 & 50.42 & 66.99 & 82.86 & 86.92 & 65.63 & 65.16 & 64.59\\
CoOp \cite{zhou2022learning} & 18.47 & 93.70 & 64.51 & 41.92 & 46.39 & 68.71 & 85.30 & 89.14 & 64.15 & 66.55 & 63.88 \\
TENT-reset \cite{wang2021tent} & 23.52 & 93.87 & 65.96 & 45.51 & 49.36 & 67.40 & 82.89 & 87.35 & 65.58 & 65.08 & 64.65 \\
TENT-continual \cite{wang2021tent} & 23.43 & 94.00 & 66.01 & 45.92 & 52.70 & 67.48 & 83.41 & 87.54 & 66.08 & 65.24 & 65.18 \\
CoTTA \cite{wang2022continual} & 14.13 & 86.98 & 35.51 & 32.39 & 14.77 & 43.77 & 32.27 & 61.71 & 41.63 & 37.64 & 40.08 \\
RoTTA \cite{yuan2023robust} & 23.43 & 94.00 & 65.91 & 45.57 & 49.53 & 67.28 & 82.94 & 87.41 & 65.61 & 65.08 & 64.68 \\
TPT \cite{shu2022test} & 24.78 & 94.16 & 66.87 & 47.75 & 42.44 & 68.98 & 84.67 & 87.79 & 65.50 & 68.04 & 65.10\\
Adaprompt \cite{wang2024towards} & 21.27 & 94.08 & 63.60 & 44.44 & 48.40 & 71.78 & 84.55 & 90.24 & - & 67.25 & -\\
TDA \cite{karmanov2024efficient} & 23.91 & 94.24 & 67.28 & 47.40 & 58.00 & 71.42 & 86.14 & 88.63 & 67.62 & 70.66 & 67.53\\
WATT \cite{osowiechi2024watt} & 24.27 & 93.27 & 66.37 & 46.39 & 55.95 & 68.21 & 83.26 & 88.09 & 65.89 & 66.01 & 65.77 \\ 
DPE \cite{zhang2024dual} & \textbf{28.95} & \textbf{94.81} & 67.31 & \textbf{54.20} & 55.79 & 75.07 & 86.17 & 91.14 & 70.07 & 70.44 & 69.40\\
DynaPrompt \cite{xiao2025dynaprompt} & 24.33 & 94.32 & 67.65 & 47.96 & 42.28 & 69.95 & 85.42 & 88.28 & 66.32 & 68.72 & 65.52\\
DOTA \cite{han2024dota} & 25.59 & 94.32 & 69.48 & 47.87 & 57.65 & 74.67 & 87.02 & 91.69 & 69.70 & \textbf{72.06} & 69.01\\
HisTPT \cite{zhang2024historical} & 26.90 & 94.50 & 69.20 	& 48.90 & 49.70 & 71.20 & \textbf{89.30} & 89.10 & 67.20 & 70.10 & 67.60 \\
BCA \cite{zhou2025bayesian}	& 28.59 & 94.69 & 66.86 & 53.49 & 56.63 & 73.12 & 85.97 & 90.43 & 68.41 & 67.59 & 68.59 \\ 
\rowcolor{grayblue}
\textbf{BayesTTA(Ours)} & 26.64 & 94.44 & \textbf{70.51} & 48.11 & \textbf{76.69} & \textbf{75.19} & 87.35 & \textbf{92.61} & \textbf{70.72} & 71.21 & \textbf{71.35} \\
\bottomrule[1pt]
\end{tabular}
\end{table*}

\textbf{Results on the cross-domain benchmark.} To evaluate the generalization ability of BayesTTA beyond continual adaptation scenarios, we assess its performance under the standard TTA setting across the cross-domain benchmark. As summarized in Table \ref{tab:cross-datasets}, BayesTTA achieves the highest average accuracy of 71.35\% with the ViT-B/16 backbone, outperforming SOTA methods including TDA, WATT, and DPE. It also surpasses recent competitive approaches, including DOTA (+2.34\%) and BCA (+2.76\%), which represent distribution modeling and Bayesian adaptation strategies, respectively. Notably, BayesTTA demonstrates strong backbone-agnostic adaptability, delivering competitive results with ResNet-50 as well, underscoring its broad applicability. Furthermore, BayesTTA demonstrates strong domain generalization, particularly in remote sensing scenarios such as EuroSAT, where it significantly surpasses TDA and DPE. While DPE slightly outperforms BayesTTA on certain individual datasets such as DTD, BayesTTA consistently ranks among the top performers across all domains, yielding the highest overall mean accuracy. Results for Adaprompt and DynaPrompt on SUN397 are marked as ``--'' due to out-of-memory errors during reproduction. Collectively, these findings validate that BayesTTA generalizes effectively under the standard TTA setting, extending its effectiveness beyond CT-TTA.

\begin{figure*}[!t]
	\centering
        \includegraphics[width=\textwidth]{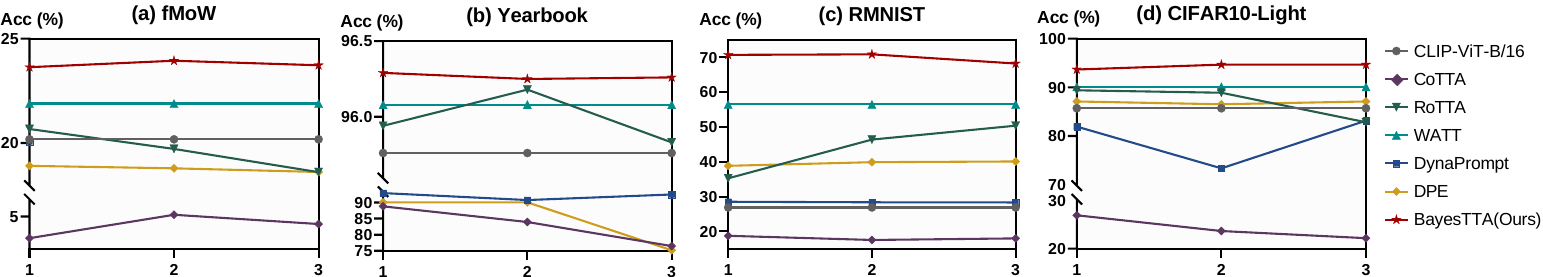}
	\caption{Performance comparisons in long-term CT-TTA adaptation. The horizontal axis represents the round number, and the vertical axis represents the average accuracy per round.}
	\label{fig:long-term}
\end{figure*}

\subsection{Comparison in the Long-Term Adaptation}
\label{subsection:long-ctta}
\vspace{-0.3\baselineskip}
We evaluate long-term adaptation under CT-TTA across three consecutive test rounds (Fig. \ref{fig:long-term}), simulating realistic deployment scenarios where models must continually adapt to temporally evolving distributions while retaining prior knowledge. On CIFAR10-Light, conventional CTTA methods like CoTTA and RoTTA suffer severe degradation due to compounding errors from erroneous pseudo-labels under continuous drift, leading to error accumulation and catastrophic forgetting. WATT alleviates this by resetting model states per batch; however, its stateless design impedes effective temporal modeling, resulting in suboptimal and static predictions. DynaPrompt partially exploits temporal context via a prompt cache but is constrained by memory limitations, as its cache is updated in a first-in-first-out manner that restricts long-range adaptation. 

In contrast, BayesTTA incrementally estimates class-conditional Gaussian distributions to capture evolving data statistics while avoiding explicit storage of past samples. EMA-based updates further stabilize adaptation over time. Its distribution-aware logit fusion combines semantic priors with calibrated predictions, enhancing robustness under non-stationary test streams. Consequently, BayesTTA consistently improves across rounds, achieving superior accuracy on fMoW, Yearbook, RMNIST, and CIFAR10-Light, effectively mitigating catastrophic forgetting and error accumulation amid evolving distribution shifts.

\subsection{Ablation Studies}
\label{subsection:ablation-studies}

\begin{table}
\centering
\caption{Ablation study of BayesTTA components on the CT-TTA benchmark (average top-1 accuracy).}
\tabcolsep=0.1cm
\begin{tabular}{clccc}
\toprule
 & Method & fMoW & RMNIST & CIFAR10-Light\\
\midrule
 & CLIP-ResNet-50 & 12.32 & 27.77 & 60.35 \\
\midrule
\rowcolor{grayblue}
(a) & BayesTTA (full) & \textbf{15.39} & \textbf{75.07} & \textbf{81.59}\\
\hdashline
(b) & \hspace{0.3em}w/o Hypothesis Test & 14.33 &  17.58 & 80.61\\
(c) & \hspace{0.3em}w/o Incremental EM & 13.28 & 53.52 & 80.26\\
(d) & \hspace{0.3em}w/o Logit Fusion & 15.11 & 74.15 & 78.52\\
(e) & \hspace{0.3em}w/o Self-paced Optimization & 15.33 & 38.58 & 65.45\\
(f) & \hspace{0.3em}w/o Continual Mechanism & 13.18 & 70.36 & 80.58\\
\bottomrule
\end{tabular}
\label{table:ablation-component}
\end{table}

\begin{figure*}[!t]
	\centering
        \includegraphics[width=0.9\textwidth]{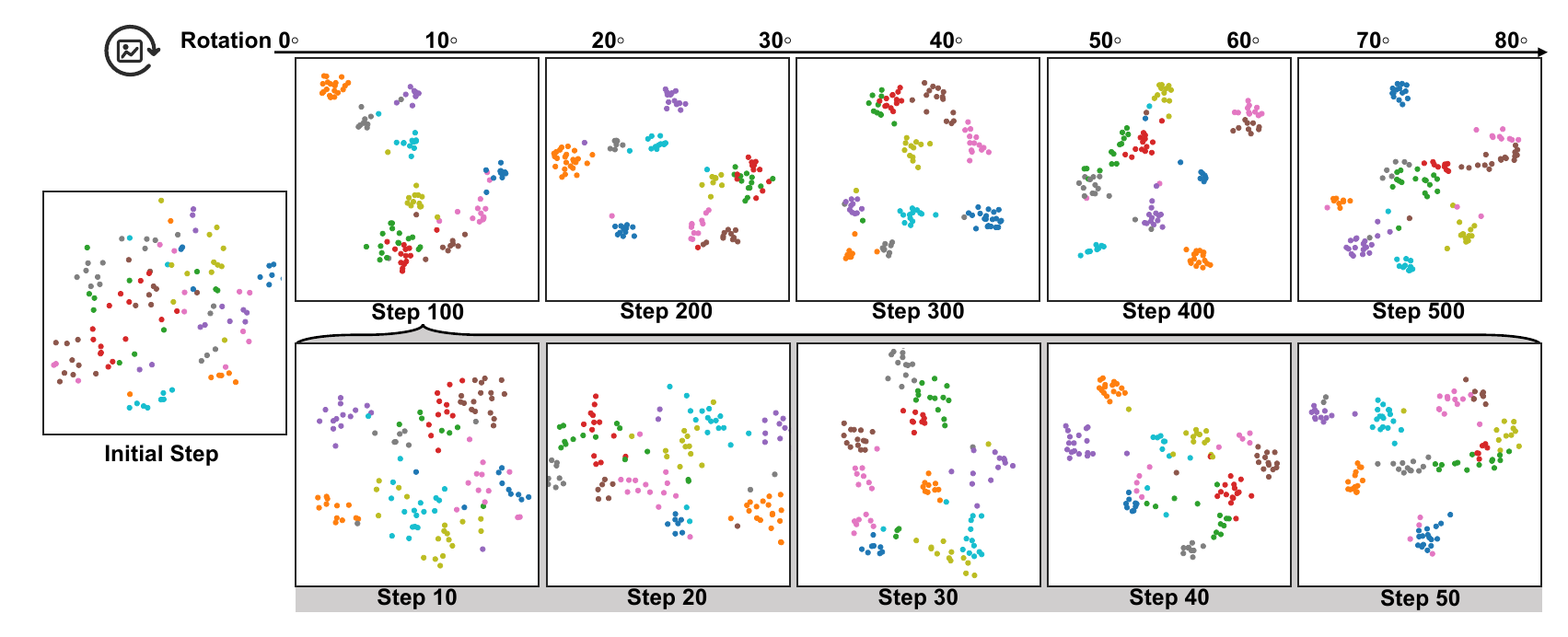}
	\caption{t-SNE visualization of BayesTTA on RMNIST across temporal test steps. Top: As digit rotation gradually evolves from $0^\circ$ to $80^\circ$, BayesTTA progressively aligns CLIP features into more compact and separable clusters. Bottom: A fine-grained view illustrates the step-wise improvement in representation structure during adaptation.}
	\label{fig:tsne}
\end{figure*}

\textbf{Ablation study of BayesTTA components.} We conduct ablation studies to assess the contribution of each component in BayesTTA. Table \ref{table:ablation-component} reports the average top-1 accuracy across all domains for each dataset. Removing any component leads to performance degradation, underscoring the necessity of their joint design. Notably, disabling the covariance homogeneity test hinders the model's ability to distinguish between homogeneous and heterogeneous covariance structures, leading to mismatched distributional assumptions and inaccurate parameter estimation—critical for reliable temporal calibration. The incremental EM mechanism facilitates the accumulation of statistical evidence over time, supporting long-range adaptation while mitigating overfitting to transient batch-specific variations. Removing logit fusion eliminates the semantic prior encoded by the pre-trained CLIP sketch, diminishing pseudo-label reliability and weakening adaptation robustness. Without self-paced optimization, the image encoder lacks calibrated supervision and remains static, failing to realign its representation space with the temporally evolving target distribution. Finally, removing the continual mechanism prevents the exploitation of previously accumulated knowledge, severely constraining long-term adaptation across gradual distribution shifts. Collectively, these findings affirm that each component plays a complementary and critical role in enhancing BayesTTA's stability and adaptability.

\textbf{t-SNE Analysis across temporal test steps.}
As shown in Fig. \ref{fig:tsne}, we qualitatively evaluate BayesTTA under continual temporal distribution shifts by visualizing the evolution of model output distributions using t-SNE \cite{van2008visualizing} on the RMNIST dataset. We use CLIP with ResNet-50 to extract features across steps 0 to 500, spanning nine gradually rotated domains from $0^\circ$ to $80^\circ$. At the initial step, the zero-shot CLIP model yields entangled and dispersed representations, exposing its susceptibility to distribution shift. As adaptation progresses (e.g., step 10 to 50), BayesTTA progressively restructures the feature space to match the evolving input stream, producing increasingly compact and separable class clusters. This transformation is accompanied by reduced inter-class confusion and sharper decision boundaries, reflecting more confident and discriminative predictions. Notably, the learned representations retain geometric coherence despite continuous domain evolution, showcasing BayesTTA's robustness under CT-TTA. These improvements stem from its principled synergy of debiased temporal distribution estimation, GDA-guided prediction calibration, and lightweight self-paced updates.

\begin{figure}[!t]
	\centering
        \includegraphics[width=0.50\textwidth]{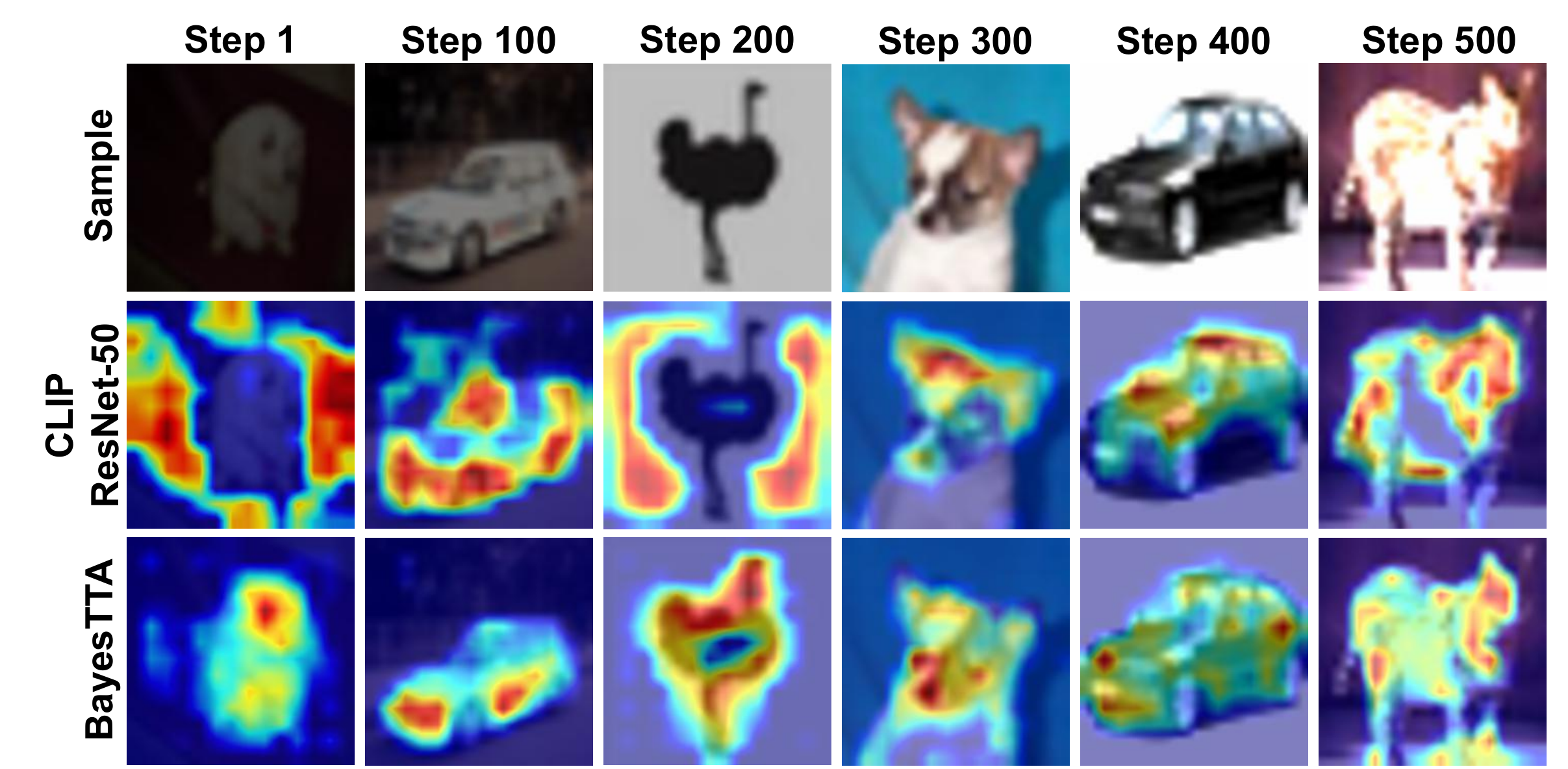}
	\caption{Comparison of Grad-CAM maps on CIFAR10-Light during adaptation. BayesTTA enhances class-discriminative focus across time, improving localization consistency.}
	\label{fig:gradcam}
\end{figure}

\textbf{Attention map analysis.}
To better understand how BayesTTA realigns spatial attention under temporally evolving distribution shifts, we visualize Grad-CAM \cite{selvaraju2017grad} on CIFAR10-Light across representative adaptation steps, covering the full illumination range from low-light to overexposure. As shown in Fig. \ref{fig:gradcam}, zero-shot CLIP exhibits diffuse and inconsistent attention, focusing on irrelevant background regions. In contrast, BayesTTA consistently enhances class-discriminative focus, yielding attention maps that align with semantically meaningful object regions. These findings demonstrate that BayesTTA achieves distribution-aware calibration and robust spatial alignment by dynamically grounding representations to task-relevant regions during continual distributional drift.

\begin{figure}[!t]
\centering
\begin{subfigure}[b]{0.48\linewidth}
    \centering
    \includegraphics[width=\linewidth]{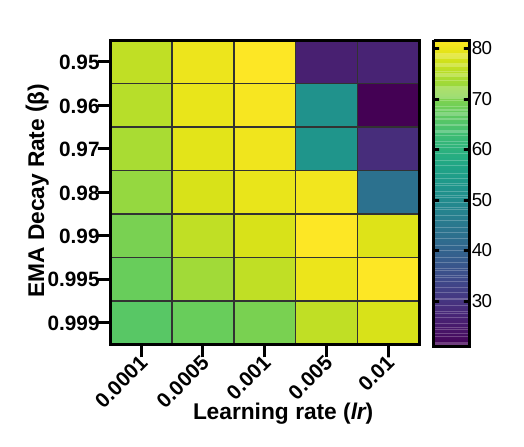}
    \caption{Sensitivity of $lr$ and $\beta$}
    \label{fig:ablation_hyper_lr_ema}
\end{subfigure}
\hfill
\begin{subfigure}[b]{0.48\linewidth}
    \centering
    \includegraphics[width=\linewidth]{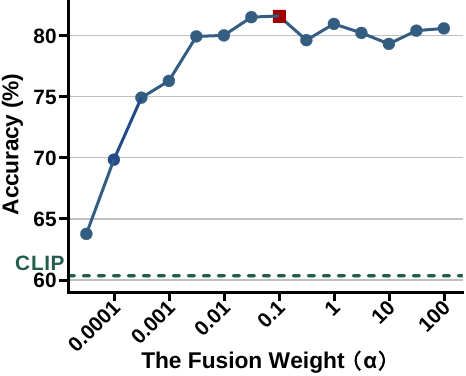}
    \caption{Sensitivity of $\alpha$}
    \label{fig:ablation_hyper_alpha}
\end{subfigure}
\caption{Hyperparameter sensitivity analysis of BayesTTA.}
\label{fig:ablation_hyper}
\end{figure}

\begin{figure}[!t]
  \centering
  \begin{minipage}{0.20\textwidth}
    \centering
    \includegraphics[width=\linewidth]{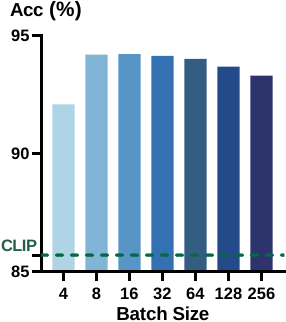}
    \caption{The impact of batch size.}
    \label{fig:ablation_batchsize}
  \end{minipage}
  \begin{minipage}{0.23\textwidth}
    \centering
    \includegraphics[width=\linewidth]{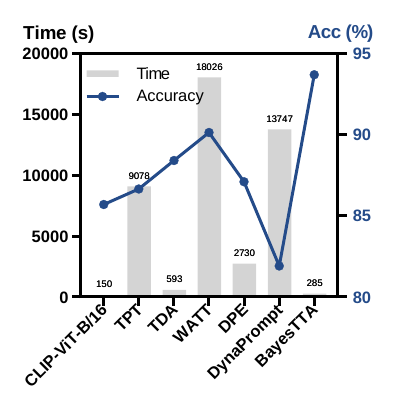}
    \caption{Computational efficiency and effectiveness.}
    \label{fig:computational}
  \end{minipage}
\end{figure}

\textbf{Hyperparameters sensitivity analysis.} We investigate the sensitivity of BayesTTA to three key hyperparameters on the CIFAR10-Light dataset: the learning rate $lr$ for self-paced optimization, the EMA decay rate $\beta$ for temporal smoothing, and the fusion weight $\alpha$ for balancing CLIP's semantic prior with distribution-aware predictions. As illustrated in Fig. \ref{fig:ablation_hyper_lr_ema} and \ref{fig:ablation_hyper_alpha}, BayesTTA maintains robust performance across a wide hyperparameter range. A moderate learning rate (e.g., $0.005$) ensures stable encoder updates, while overly small or large values may lead to underfitting or training instability. The EMA decay $\beta$ controls how strongly the model incorporates historical knowledge; lower values risk overfitting to transient test batches, whereas higher values slow down adaptation. The fusion weight $\alpha$ governs the trade-off between CLIP's semantic prior and GDA-based calibration. Over-reliance on either side compromises alignment—insufficient fusion weakens semantic grounding, whereas excessive dependence on GDA amplifies early noise. Overall, BayesTTA exhibits resilience to hyperparameter fluctuations and benefits from well-balanced configurations that jointly promote stability, adaptability, and semantic alignment.

\textbf{Model performance across batch sizes.} We analyze the sensitivity of BayesTTA to different batch sizes under the CIFAR10-Light dataset. As shown in Fig. \ref{fig:ablation_batchsize}, performance generally improves with larger batches, while remaining competitive even under small-batch settings. For example, with a batch size of 8, BayesTTA achieves 92.26\% accuracy, notably surpassing the zero-shot CLIP-ViT-B/16 baseline of 85.70\%. To balance adaptation efficiency and memory consumption, we adopt a batch size of 128 in experiments, consistent with SOTA methods \cite{osowiechi2024watt}.

\subsection{Computational Complexity}
\label{subsection:computational-complexity}
We evaluate the computational efficiency of BayesTTA compared to CLIP-ViT-B/16 and SOTA CTTA methods for VLMs on the CIFAR10-Light dataset (70,000 images). All experiments are conducted on a single NVIDIA RTX 3090 GPU. As shown in Fig. \ref{fig:computational}, we report total inference time (including data loading) and average accuracy. For a fair comparison, CLIP, WATT, and BayesTTA use a batch size of 128, while TPT, TDA, DPE, and DynaPrompt are limited to a batch size of 1 due to algorithmic constraints that prevent efficient batch-wise processing.

BayesTTA achieves 93.68\% accuracy in 285 seconds, outperforming CLIP-ViT-B/16 (85.7\%, 150 s) with only a moderate 135-second overhead. Compared to TPT (86.65\%, 9078 s), BayesTTA yields a +7.03\% accuracy improvement while running approximately 32× faster, as TPT suffers from high latency due to multi-view augmentation and repeated forward passes. BayesTTA also surpasses WATT (+3.55\%, 63× faster), which incurs substantial cost from iterative ensemble-based optimization. Against DynaPrompt (81.92\%, 13747 s), BayesTTA achieves a +11.76\% gain with a 48× speedup. Additionally, BayesTTA improves over TDA (88.4\%, 593 s) and DPE (87.1\%, 2730 s), both limited by their single-sample adaptation designs. In contrast, BayesTTA enables efficient batch-wise processing by confining updates to lightweight normalization parameters within the image encoder, thereby avoiding costly full-model tuning. This design facilitates fast and scalable adaptation, highlighting BayesTTA's practicality for real-world deployment under evolving test distributions.

\section{Conclusion}
This paper introduces Continual-Temporal Test-Time Adaptation (CT-TTA), a realistic yet underexplored paradigm where test distributions evolve gradually over time, presenting unique challenges such as error accumulation, catastrophic forgetting, and representation drift. To tackle these challenges, we propose BayesTTA, a principled Bayesian adaptation framework that models evolving distributions via incremental EM, adaptively selects covariance structures through statistical hypothesis testing, and performs calibrated inference with GDA. Efficient and stable representation alignment is ensured through lightweight adaptation of normalization layers guided by fused pseudo-labels. Extensive experiments on both CT-TTA and standard TTA benchmarks demonstrate that BayesTTA consistently outperforms prior methods in accuracy and stability while maintaining high computational efficiency. We hope this work inspires further research on temporally aware test-time adaptation for the reliable deployment of VLMs in real-world scenarios.

\textbf{Limitations.} While BayesTTA effectively adapts to temporally evolving distribution shifts, two limitations remain. First, it relies on hyperparameter tuning, a common challenge among TTA methods. Second, its gradient-based updates introduce a moderate inference overhead compared to zero-shot CLIP, though the accuracy gains justify the cost. Notably, BayesTTA achieves higher efficiency than existing TTA methods, owing to its lightweight, batch-wise adaptation strategy. These trade-offs suggest potential directions for enhancing efficiency and robustness in future work.

\section*{Acknowledgments}
The authors would like to thank the associate editor and anonymous reviewers for their valuable comments. This work is supported by National Natural Science Foundation of China, Grant No. 62406313, Postdoctoral Fellowship Program, Grant No. GZC20232812, China Postdoctoral Science Foundation, Grant No. 2024M753356, 2023 Special Research Assistant Grant Project of the Chinese Academy of Sciences.

\bibliographystyle{IEEEtran}
\bibliography{IEEEabrv,reference}

\begin{IEEEbiography}[{\includegraphics[width=1in,height=1.25in,clip,keepaspectratio]{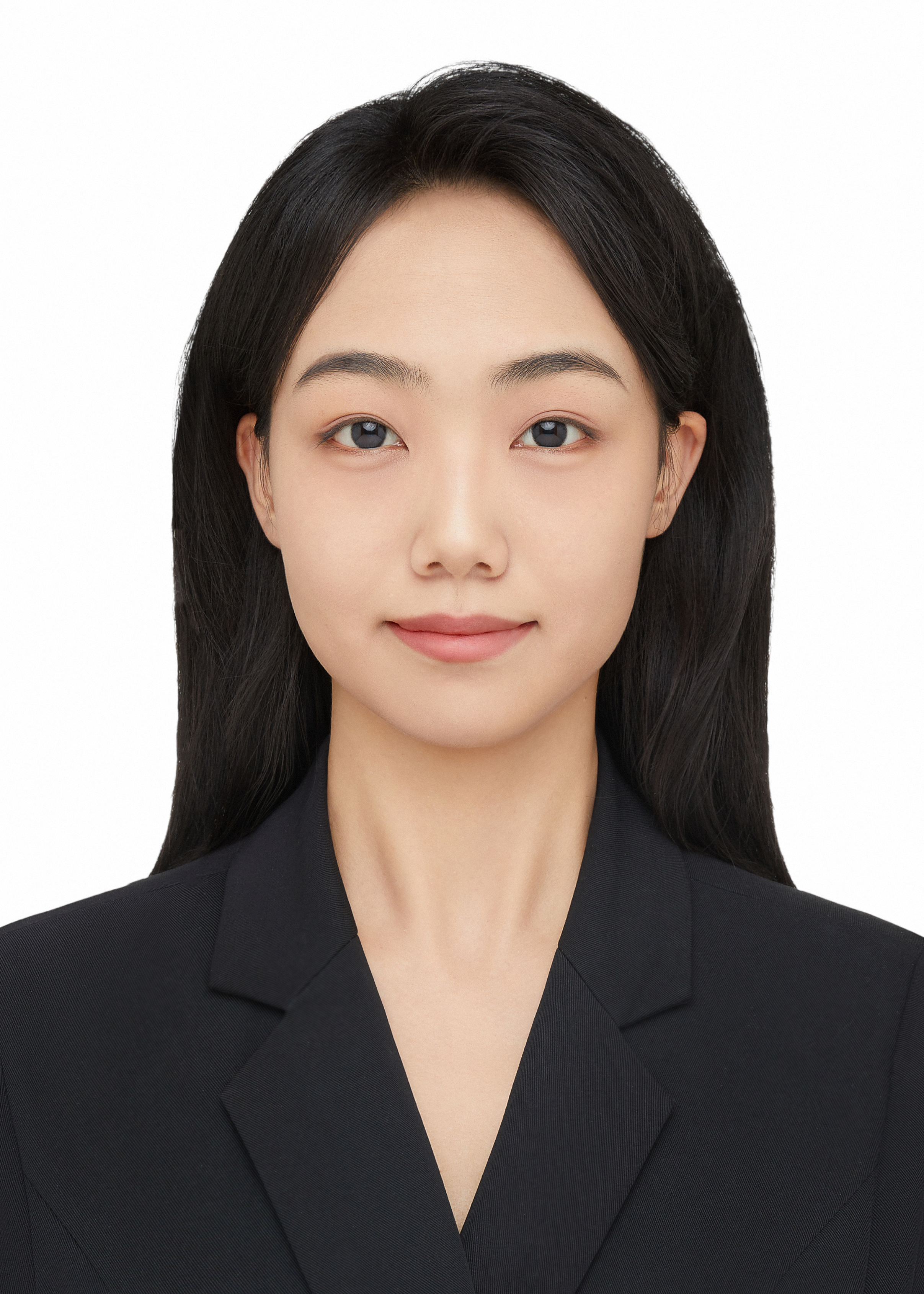}}]{Shuang Cui}
received the B.S. degree from Northwest A\&F University, Xianyang, China, in 2021. She is currently pursuing the Ph.D. degree at the University of Chinese Academy of Sciences and the Institute of Software, Chinese Academy of Sciences, Beijing, China. Her research interests include transfer learning, test-time adaptation, and continual learning.
\end{IEEEbiography}

\begin{IEEEbiography}[{\includegraphics[width=1in,height=1.25in,clip,keepaspectratio]{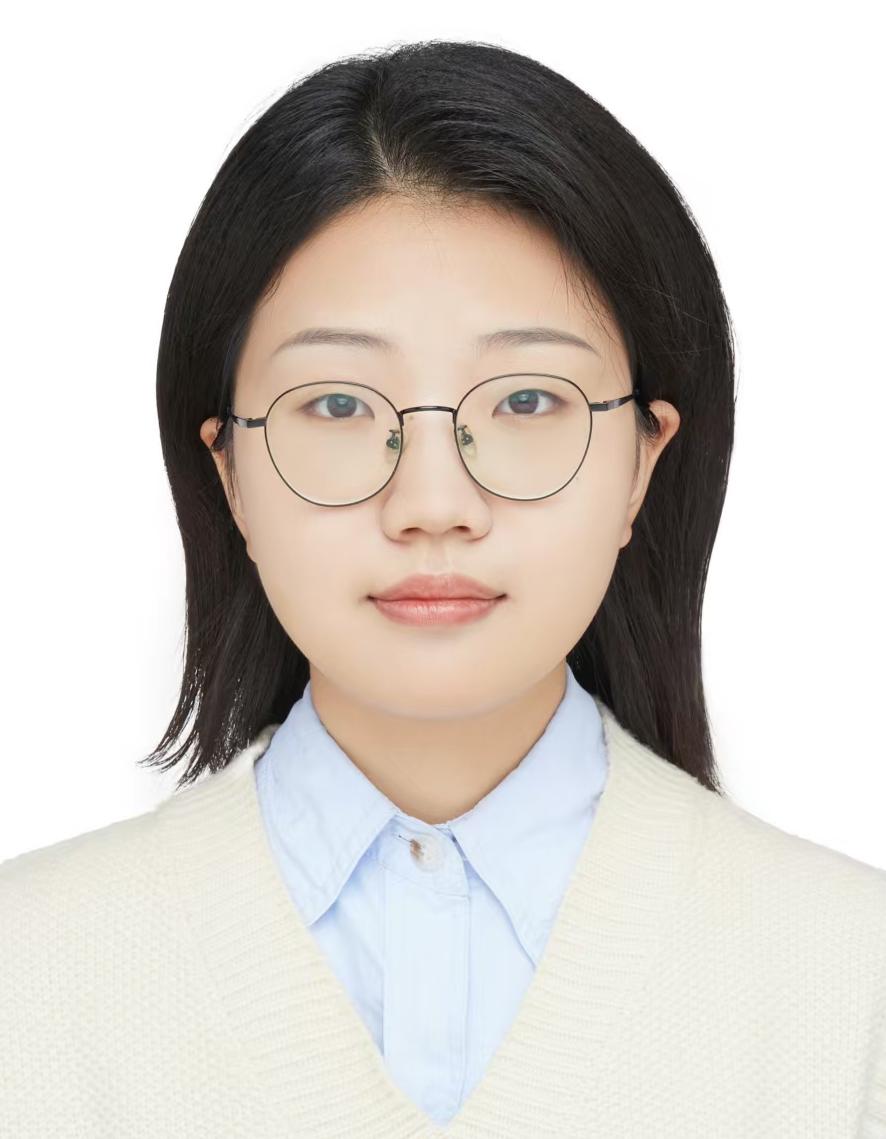}}]{Jinglin Xu}
received the B.S. degree from Beihang University, Beijing, China, in 2023. She is currently pursuing the Ph.D. degree at the University of Chinese Academy of Sciences and the Institute of Software, Chinese Academy of Sciences, Beijing, China. Her research interests include multimodal representation learning, self-supervised learning, and test-time adaptation.
\end{IEEEbiography}

\begin{IEEEbiography}[{\includegraphics[width=1in,height=1.25in,clip,keepaspectratio]{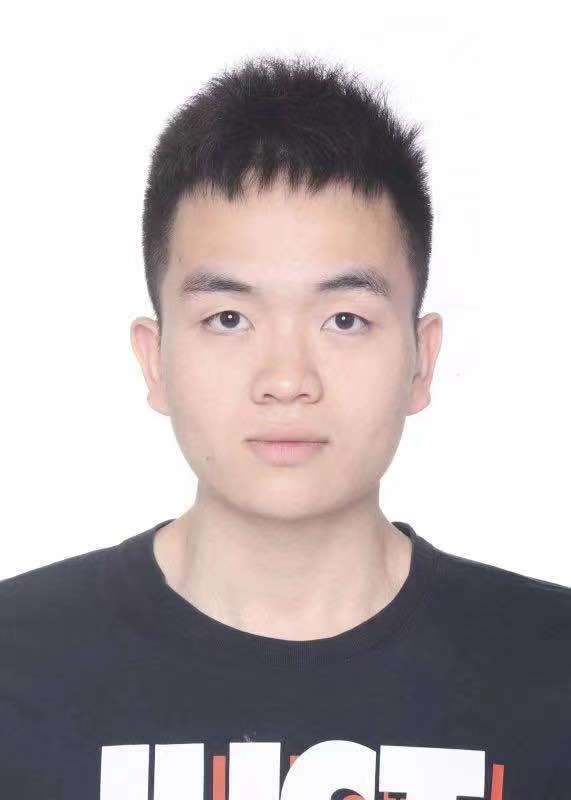}}]{Yi Li} received the B.Sc degree in the School of Fuzhou University, Fuzhou, China. He is currently working toward the Ph.D degree in University of Chinese Academy of Sciences, Beijing, China. His research interests lie in the domains of multimodal learning and causality, with a particular focus on vision-language representation learning and causal representation learning.
\end{IEEEbiography}

\begin{IEEEbiography}[{\includegraphics[width=1in,height=1.25in,clip,keepaspectratio]{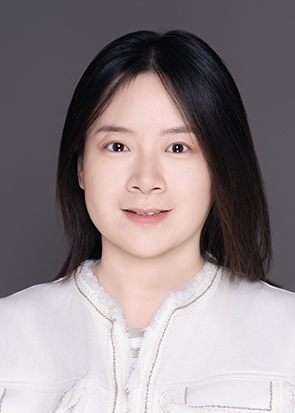}}]{Xiongxin Tang}
received her Ph.D. degree from the University of Chinese Academy of Sciences, in 2013. She currently holds the position of vice professor at the Institute of Software, Chinese Academy of Sciences. With extensive experience in optical simulation calculations and optical software development, she has actively engaged in research and development in this field. She has made significant contributions to her field with over 20 published academic papers in reputable domestic and international journals and conferences. 
\end{IEEEbiography}

\begin{IEEEbiography}[{\includegraphics[width=1in,height=1.25in,clip,keepaspectratio]{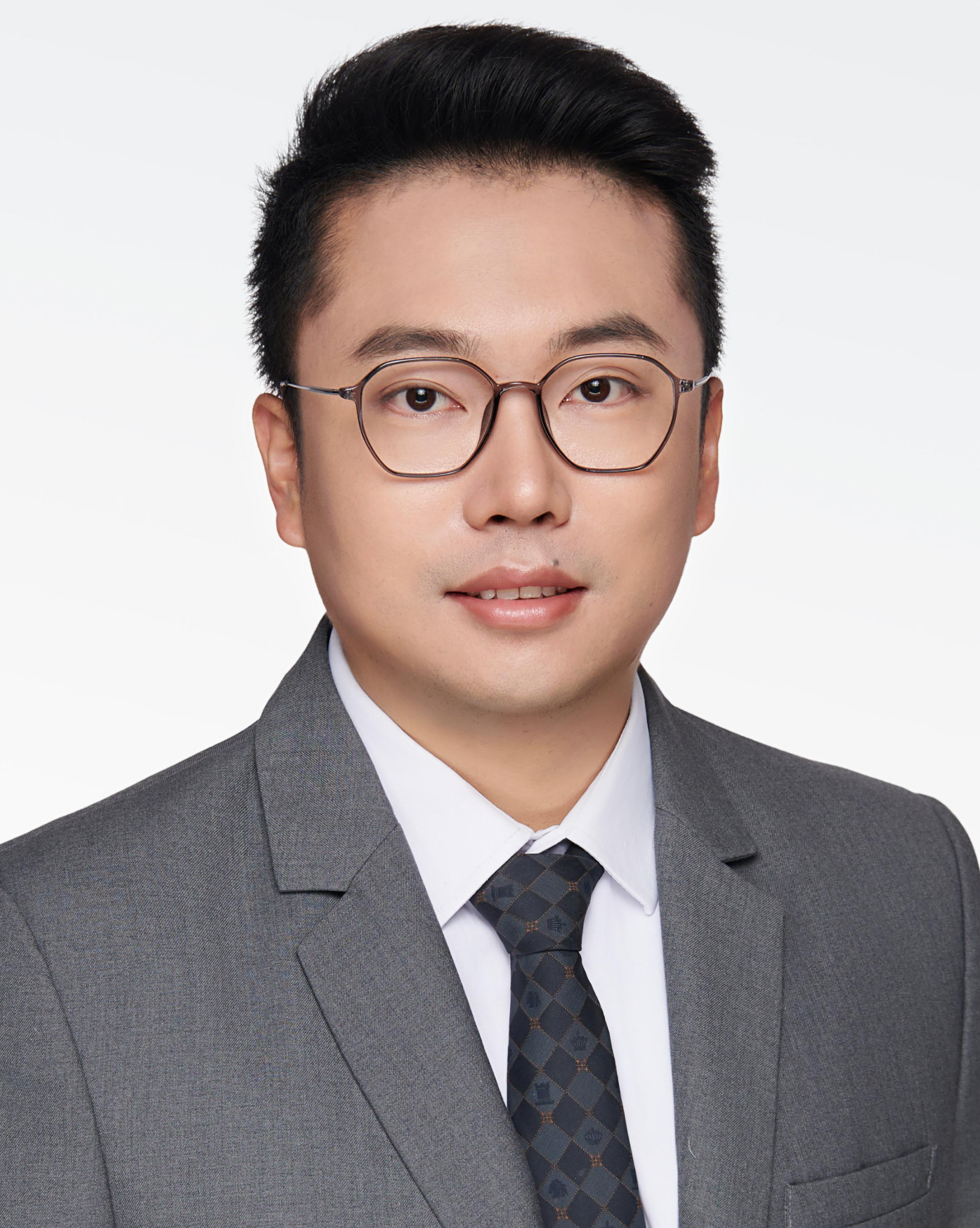}}]{Jiangmeng Li}
received the MS degree from New York University, New York, USA, in 2018, and the Ph.D degree in the Institute of Software, Chinese Academy of Sciences, Beijing, China, in 2023. He is currently an Assistant Professor at the Institute of Software, Chinese Academy of Sciences. His research interests include causal inference, self-supervised learning, and machine learning. He has published more than 40 papers in journals and conferences such as IEEE Transactions on Knowledge and Data Engineering (TKDE), International Journal of Computer Vision (IJCV), International Conference on Machine Learning (ICML), Conference and Workshop on Neural Information Processing Systems (NeurIPS), International Conference on Learning Representations (ICLR), etc.
\end{IEEEbiography}

\begin{IEEEbiography}[{\includegraphics[width=1in,height=1.25in,clip,keepaspectratio]{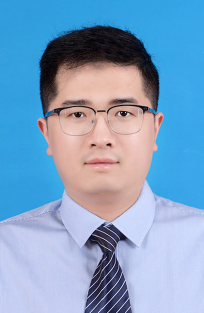}}]{Jiahuan Zhou}
received his B.E. (2013) from Tsinghua University, and his Ph.D. degree (2018) in the Department of Electrical Engineering \& Computer Science, Northwestern University. Currently, he is a Tenure-Track Assistant Professor at the Wangxuan Institute of Computer Technology, Peking University. His research interests include computer vision, deep learning, and machine learning. He has authored more than 60 papers in international journals and conferences including Nature Communications, Nature Synthesis, IEEE TPAMI, IJCV, IEEE TIP, IEEE TIFS, CVPR, ICCV, ICML, ECCV, ICLR, AAAI, IJCAI, ACM MM, and so on. He serves as an Area Chair for CVPR, ICML, NeurIPS, ICME, and ICPR, an Associate Editor of the Springer Journal of Machine Vision and Applications (MVA), a regular reviewer member for several journals and conferences, e.g., T-PAMI, IJCV, TIP, CVPR, ICCV, ECCV, NeurIPS, ICML, ICLR, AAAI, and so on.
\end{IEEEbiography}

\begin{IEEEbiography}[{\includegraphics[width=1in,height=1.25in,clip,keepaspectratio]{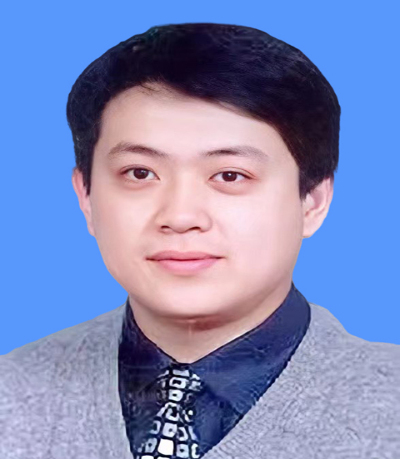}}]{Fanjiang Xu}
received his B.Sc. and M.Sc. degrees in computer science both from National University of Defense Technology and his Ph.D. degree in computer science from Huazhong University of Science and Technology. He is currently a Professor at the Institute of Software, Chinese Academy of Sciences. His research interests include intelligent information processing and integration.
\end{IEEEbiography}

\begin{IEEEbiography}[{\includegraphics[width=1in,height=1.25in,clip,keepaspectratio]{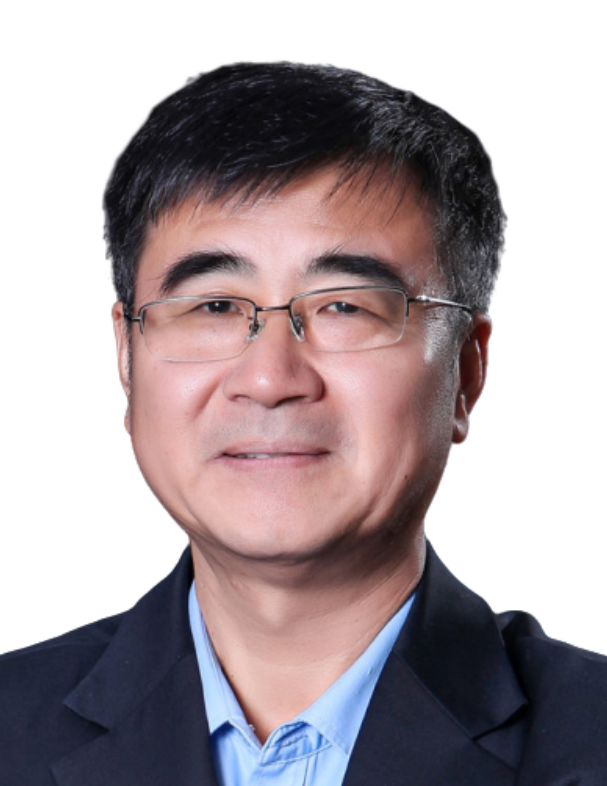}}]{Fuchun Sun} received his Ph.D. degree in computer science and technology from Tsinghua University, Beijing, China, in 1997. He is currently a Professor with the Department of Computer Science and Technology and President of Academic Committee of the Department, Tsinghua University, deputy director of State Key Lab. of Intelligent Technology and Systems, Beijing, China. His research interests include artificial intelligence, intelligent control and robotics, information sensing and processing in artificial cognitive systems, etc. He was recognized as a Distinguished Young Scholar in 2006 by the Natural Science Foundation of China. He became a member of the Technical Committee on Intelligent Control of the IEEE Control System Society in 2006. He serves as Editor-in-Chief of International Journal on Cognitive Computation and Systems, and an Associate Editor for a series of international journals including the IEEE TRANSACTIONS ON COGNITIVE AND DEVELOPMENTAL SYSTEMS, the IEEE TRANSACTIONS ON FUZZY SYSTEMS, and the IEEE TRANSACTIONS ON SYSTEMS, MAN, AND CYBERNETICS: SYSTEMS.
\end{IEEEbiography}

\begin{IEEEbiography}[{\includegraphics[width=1in,height=1.25in,clip,keepaspectratio]{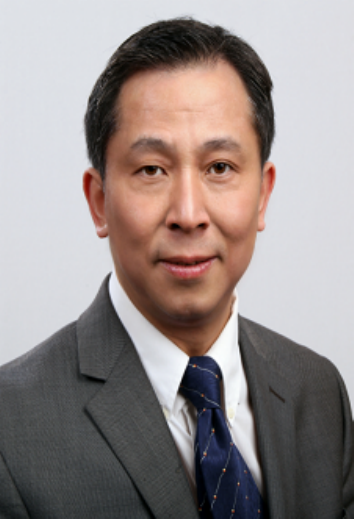}}]{Hui Xiong}
received his Ph.D. in Computer Science from the University of Minnesota - Twin Cities, USA, in 2005, the B.E. degree in Automation from the University of Science and Technology of China (USTC), Hefei, China, and the M.S. degree in Computer Science from the National University of Singapore (NUS), Singapore. He is a chair professor at the Hong Kong University of Science and Technology (Guangzhou). He is also a Distinguished Professor at Rutgers, the State University of New Jersey, where he received the 2018 Ram Charan Management Practice Award as the Grand Prix winner from the Harvard Business Review, RBS Dean's Research Professorship (2016), two-year early promotion/tenure (2009), the Rutgers University Board of Trustees Research Fellowship for Scholarly Excellence (2009), the ICDM-2011 Best Research Paper Award (2011), the Junior Faculty Teaching Excellence Award (2007), Dean's Award for Meritorious Research (2010, 2011, 2013, 2015) at Rutgers Business School, the 2017 IEEE ICDM Outstanding Service Award (2017), and the AAAI-2021 Best Paper Award (2021). Dr. Xiong is also a Distinguished Guest Professor (Grand Master Chair Professor) at the University of Science and Technology of China (USTC). For his outstanding contributions to data mining and mobile computing, he was elected an ACM Distinguished Scientist in 2014, an IEEE Fellow and an AAAS Fellow in 2020. His general area of research is data and knowledge engineering, with a focus on developing effective and efficient data analysis techniques for emerging data intensive applications. He currently serves as a co-Editor-in-Chief of Encyclopedia of GIS (Springer) and an Associate Editor of IEEE Transactions on Data and Knowledge Engineering (TKDE), IEEE Transactions on Big Data (TBD), ACM Transactions on Knowledge Discovery from Data (TKDD) and ACM Transactions on Management Information Systems (TMIS). He has served regularly on the organization and program committees of numerous conferences, including as a Program Co-Chair of the Industrial and Government Track for the 18th ACM SIGKDD International Conference on Knowledge Discovery and Data Mining (KDD), a Program Co-Chair for the IEEE 2013 International Conference on Data Mining (ICDM), a General Co-Chair for the IEEE 2015 International Conference on Data Mining (ICDM), and a Program Co-Chair of the Research Track for the 24th ACM SIGKDD International Conference on Knowledge Discovery and Data Mining (KDD2018).
\end{IEEEbiography}

\end{document}